%% file: iclr2027_conference.tex
\documentclass{article} %
\usepackage{iclr2027_conference,times}

\input{math_commands.tex}

\usepackage{hyperref}
\usepackage{url}
\usepackage{graphicx}
\usepackage{comment} %
\usepackage{booktabs} %
\usepackage{multirow}
\usepackage{tabularx,array} %
\usepackage{textcomp} %
\usepackage{amssymb}  %
\usepackage{kotex}    %
\usepackage{xspace}

\usepackage{xcolor}
\definecolor{mylink}{rgb}{0.10,0.20,0.55}
\hypersetup{colorlinks=true, allcolors=mylink}
\usepackage{enumitem} %
\usepackage[most]{tcolorbox} %
\newtcolorbox{promptbox}[2][]{enhanced, breakable, colback=black!3, colframe=black!40,
  boxrule=0.4pt, arc=2pt, left=7pt, right=7pt, top=5pt, bottom=5pt,
  fonttitle=\bfseries\small\sffamily, coltitle=black, colbacktitle=black!9,
  title={#2}, fontupper=\scriptsize\ttfamily, halign=flush left,
  before skip=6pt, after skip=8pt, #1}
\newcommand{\pfield}[1]{\par\medskip\noindent{\footnotesize\bfseries\scshape #1}\par\nopagebreak\noindent{}}

\newcommand{\method}{OmniShowcase\xspace}
\newcommand{\cmark}{\textcolor{green!60!black}{\checkmark}}          %
\newcommand{\xmark}{\textcolor{red!80!black}{\ensuremath{\times}}}

\definecolor{os00}{HTML}{774AC6}
\definecolor{os01}{HTML}{7556D0}
\definecolor{os02}{HTML}{7165DA}
\definecolor{os03}{HTML}{6775DF}
\definecolor{os04}{HTML}{537ADE}
\definecolor{os05}{HTML}{3C7DE2}
\definecolor{os06}{HTML}{3187E2}
\definecolor{os07}{HTML}{5B9BCB}
\definecolor{os08}{HTML}{B3A262}
\definecolor{os09}{HTML}{C59711}
\definecolor{os10}{HTML}{CF9100}
\definecolor{os11}{HTML}{DE9100}
\newcommand{\methodgrad}{%
  \textcolor{os00}{O}%
  \textcolor{os01}{m}%
  \textcolor{os02}{n}%
  \textcolor{os03}{i}%
  \textcolor{os04}{S}%
  \textcolor{os05}{h}%
  \textcolor{os06}{o}%
  \textcolor{os07}{w}%
  \textcolor{os08}{c}%
  \textcolor{os09}{a}%
  \textcolor{os10}{s}%
  \textcolor{os11}{e}%
}

\title{\texorpdfstring{\methodgrad}{OmniShowcase}: Omnidirectional Subject-Consistent Video Generation from Multi-View Images}

\newcommand{\authorbreak}{\end{tabular}\hfil\linebreak[4]\hfil
  \begin{tabular}[t]{l}\bf\ignorespaces}

\author{Junyeong Ahn$^{1*}$, \; Geonwoo Kim$^{2*}$, \; Kinam Kim$^{1*}$, \; Gihwi Kim$^{3}$,
\authorbreak
Hoonjin Jung$^{3}$, \; Hyojin Jang$^{1}$, \; Jaegul Choo$^{1}$ \\
$^{1}$KAIST AI \quad $^{2}$Inha University \quad $^{3}$FLIPTION \\
}

\iclrfinalcopy %
\begin{document}

\maketitle

\begin{figure}[!h]
    \centering
    \vspace{-9mm}
    \includegraphics[width=\linewidth]{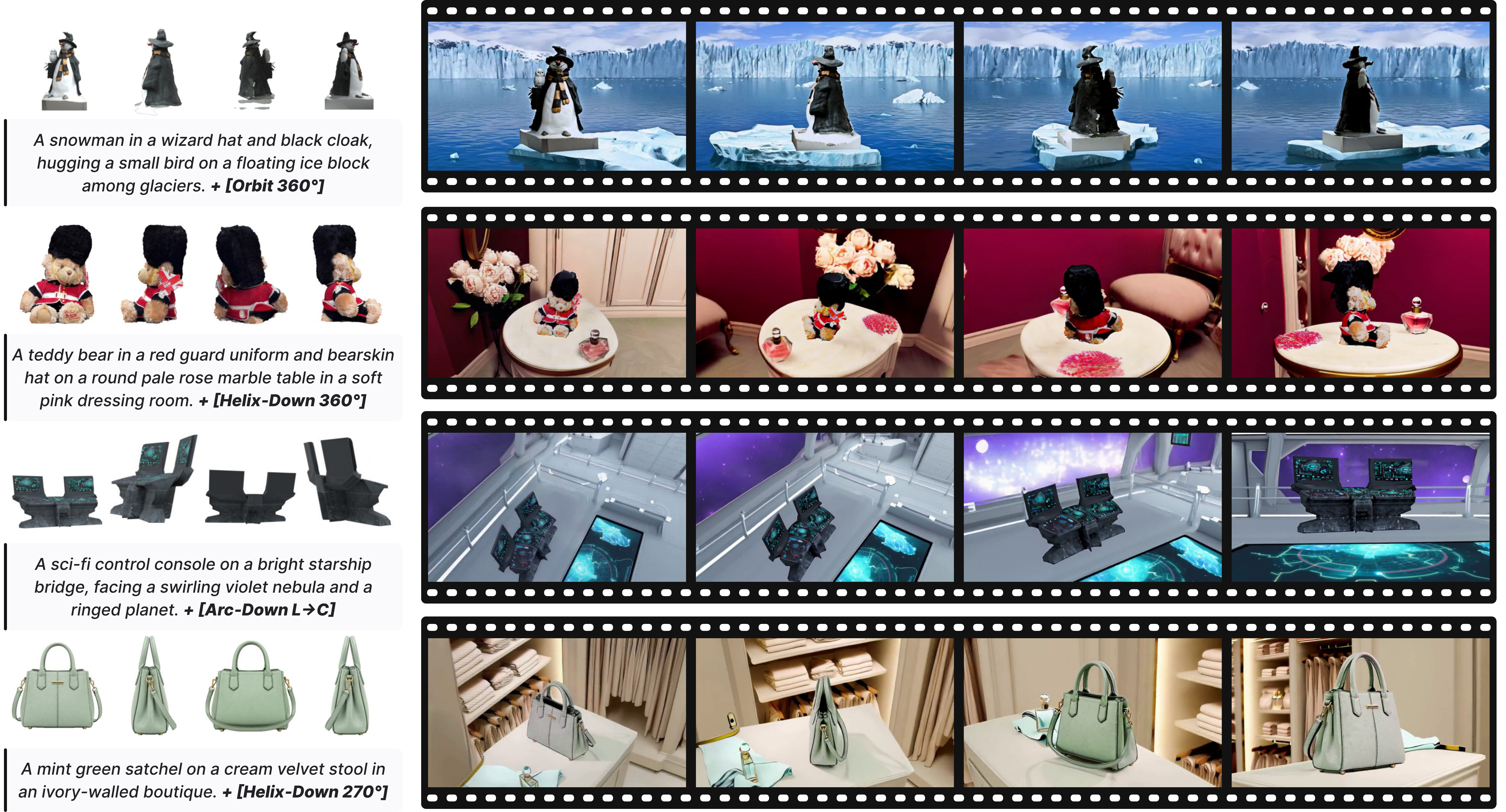}
    \vspace{-0.6em}
    \caption{\textbf{\method} produces subject-driven videos that stay
    3D-consistent across every revealed viewpoint.}
    \label{fig:teaser}
\end{figure}

\input{sec/0_abstract}

\input{sec/1_intro}
\input{sec/2_related_work}

\input{sec/3_method}
\input{sec/4_experiment}
\input{sec/5_conclusion}

\clearpage
\subsection*{AI use statement}

In the preparation of this work, we used LLMs to polish the text, including
grammar checking and phrase refinement, and to polish figure illustrations.
Generative models are also components of the data pipeline itself, where a
vision-language model gates asset and clip curation, a large language model
writes the scene vocabulary, and a video-language model captions each clip.
Appendices~\ref{app:dataset} and~\ref{app:rubric} state where each one is used
and give every prompt and rubric verbatim. No LLM contributed to the research
ideas, the method design or the analysis.

We take responsibility for the final content of this work, including text, claims or artifacts produced with the aid of
generative AI.

\subsection*{Ethics statement}

This work includes a human preference study with $44$ participants, as described
in Appendix~\ref{app:userstudy}. Every 3D asset comes
from Objaverse, the Digital Twin Catalog or Google Scanned Objects under a
Creative Commons license, and none carries a NoDerivatives term, so the renders
may be released. A small fraction is NonCommercial, which the research-only
release respects. We ship the license and the original author of every asset
with the renders, and ShareAlike terms carry over to the renders that inherit
them. An asset whose terms the release cannot meet is replaced by a pointer to
the original and the code that reproduces our renders from it.

\subsection*{Reproducibility statement}

We release the sources of the data generation, the exact modules composing the
pipeline, the method architecture, the VLM and LLM rubrics used for curation,
and the prompts that generate the captions, all in
Appendices~\ref{app:dataset} and~\ref{app:rubric}. The evaluation protocol is
fixed as described in Section~\ref{subsec:setup}, and where a method requires a
deviation we state it, such as extending Phantom and MAGREF from single-view to
multi-view input and appending the camera-trajectory sentence that our model
consumes. The data-generation code and the dataset will be released publicly.

\clearpage
\bibliography{iclr2027_conference,main}
\bibliographystyle{iclr2027_conference}

\appendix
\input{sec/X_suppl}

\end{document}

%% file: math_commands.tex
\usepackage{amsmath,amsfonts,bm}

\def\eqref#1{equation~\ref{#1}}

\def\1{\bm{1}}

\DeclareMathAlphabet{\mathsfit}{\encodingdefault}{\sfdefault}{m}{sl}
\SetMathAlphabet{\mathsfit}{bold}{\encodingdefault}{\sfdefault}{bx}{n}

%% file: sec/0_abstract.tex
\begin{abstract}

Generating videos of a specific subject from a few reference images remains challenging when the camera reveals surfaces that are not visible in the input.
Early subject-to-video (S2V) methods condition on a single reference image and must therefore infer the subject's unseen appearance and geometry as the viewpoint changes. To address this limitation, recent methods extend S2V to multiple reference views, providing more information about the subject from different viewpoints. However, they do not explicitly specify the 3D structure that connects these observations. In addition, existing training data do not guarantee that viewpoint changes preserve the same underlying 3D subject. We present \method, built on two components: (i) a \emph{fully automated} data synthesis pipeline that produces training videos  along diverse camera trajectories, with viewpoint variation driven by underlying subject geometry, even spanning full 360-degree orbits; and (ii) a geometry-conditioning method that provides the video model with a turntable of surface-normal maps reconstructed from the reference views without requiring temporal alignment with the generated frames. We construct \textsc{Showcase-4K}, a corpus of 3D-asset-based videos covering twelve camera motions across $88$ trajectory configurations with reconstruction-derived surface-normal conditioning, and \textsc{ShowcaseBench}, a held-out evaluation set with the source meshes and camera parameters used to render each video.
Experiments show state-of-the-art subject identity and 3D consistency while maintaining competitive video quality and motion. Project page: \url{justin4ai.github.io/OmniShowcase}

\end{abstract}

%% file: sec/1_intro.tex
\section{Introduction}
\label{sec:intro}

Subject-to-video (S2V) generation aims to create videos of a specified subject from reference images while preserving its appearance across time and viewpoints.
Recent video diffusion models generate photorealistic videos from text or a single image~\citep{yang2025cogvideox,kong2024hunyuanvideo,wan2025,brooks2024sora}, and S2V methods use reference images to preserve subject identity in generated videos~\citep{liu2025phantom,deng2025magref,jiang2025vace,yuan2025consisid}.
However, most S2V methods condition on a single reference image, which provides appearance information only for the visible surfaces and leaves other regions unconstrained.
As the camera reveals previously unseen sides, generated videos may exhibit inconsistent geometry and appearance, which can lead to changes in silhouette, structure, and surface details.
These ambiguities motivate the use of multiple reference views and explicit geometric information. The main difficulty is geometric rather than purely photometric since the model must infer 3D structure that is absent from the input.

Recent methods incorporate multiple reference views into S2V generation.
MV-S2V~\citep{song2026mvs2v} trains a feed-forward model using generated multi-view videos, whereas 3DreamBooth~\citep{ko2026threedreambooth} performs per-subject optimization using static multi-view images.
These efforts motivate two complementary directions: training on geometrically consistent viewpoint changes and providing explicit 3D geometry during generation. For the latter, recent feed-forward reconstruction methods can recover 3D geometry from a few uncalibrated images~\citep{wang2024dust3r,wang2025vggt,wang2025pi3,lin2025depthanything3}, making explicit geometry available from the same sparse references used at inference.

We present \textbf{\method} (Figure~\ref{fig:teaser}), a framework that combines an automated data synthesis pipeline with geometry-aware conditioning.
The pipeline renders multi-view videos from 3D assets and composites them into realistic scenes.
For conditioning, we reconstruct geometry from the reference views and render a surface-normal sequence using the same procedure during training and inference.
Our contributions are as follows:
\begin{itemize}[topsep=3pt, itemsep=2pt, parsep=0pt]
    \item We introduce an automated data synthesis pipeline that renders subject views from 3D assets and composites them into realistic scenes to construct training videos.
    \item We present \textsc{Showcase-4K}, a dataset of 4,000 videos paired with diverse source meshes, camera trajectories, reference images, and surface-normal sequences, along with \textsc{ShowcaseBench}, a held-out benchmark with ground-truth geometry and camera parameters for evaluating subject consistency across viewpoints. We will publicly release \textsc{Showcase-4K}, \textsc{ShowcaseBench}, and the associated data-synthesis code.
    \item We propose a geometry-aware in-context conditioning mechanism that uses reference images and a reconstructed surface-normal sequence to guide video generation without requiring frame-wise alignment between the normal and target sequences.
    \item We evaluate our method against single- and multi-view S2V baselines on synthetic and real-world subjects, achieving state-of-the-art performance in subject identity preservation and 3D consistency while maintaining competitive video quality.
\end{itemize}

%% file: sec/2_related_work.tex
\section{Related Work}
\label{sec:related_work}

\subsection{Video Generation Diffusion Models}
\label{subsec:rw_video}

Diffusion models~\citep{ho2020ddpm,rombach2022ldm} are the dominant approach to visual content synthesis, and the shift from convolutional U-Nets to transformer denoisers~\citep{peebles2023dit} with rectified-flow training~\citep{esser2024sd3} improved their scalability and sample quality. These advances carried over to video. Early video models extended pretrained image backbones with temporal layers~\citep{singer2023makeavideo,blattmann2023svd,guo2024animatediff}; more recent systems use large diffusion transformers trained on web-scale corpora and generate high-resolution clips of steadily increasing length~\citep{yang2025cogvideox,kong2024hunyuanvideo,wan2025,brooks2024sora}. Image-to-video (I2V) variants add a starting frame as a condition, giving coarse control over appearance. Our framework builds on a pretrained I2V diffusion transformer and adapts it through low-rank tuning~\citep{hu2022lora}.

\subsection{3D Generation and Novel View Synthesis}
\label{subsec:rw_3d}

Representing and rendering 3D content from images has been studied through neural fields~\citep{mildenhall2020nerf} and point-based splatting~\citep{kerbl20233dgs}. To synthesize unseen viewpoints without per-scene reconstruction, one line of work distills 2D diffusion priors into multi-view-consistent generators: image-conditioned models such as \citet{liu2023zero123} hallucinate novel views of an object, and multi-view diffusion models~\citep{shi2023mvdream,liu2024syncdreamer,voleti2024sv3d} jointly denoise several views to enforce cross-view coherence. These methods are mostly confined to isolated objects against clean backgrounds and do not model scene-level dynamics. In parallel, feed-forward geometry estimators build on monocular depth foundation models~\citep{yang2024depthanything} and recover dense 3D structure from sparse, uncalibrated images in a single pass~\citep{wang2024dust3r,wang2025vggt,wang2025pi3,lin2025depthanything3}. This makes it practical to derive explicit geometry from multi-view reference images. We use this capability to supplement coarse 3D information of unseen views as a geometry cue.

\begin{table}[!t]
\centering
\small
\setlength{\tabcolsep}{4pt}
\renewcommand{\arraystretch}{1.15}
\caption{\textbf{Comparison of subject-to-video methods.}
\emph{Data}: representative
training-set order of magnitude. \emph{General.}: generalization without optimization for inference.
Geom.: explicit geometric conditioning. \emph{Supervision}: source of the
training signal.}
\label{tab:requirements}
\begin{tabular}{lccccl}
\toprule
Method & Data & References & General. & Geom. & Supervision \\
\midrule
Phantom~\citep{liu2025phantom}            & 1M    & single view & \cmark & \xmark & real video \\
MAGREF~\citep{deng2025magref}             & large & single view & \cmark & \xmark & real video \\
MV-S2V~\citep{song2026mvs2v}              & 14K   & multi-view (4) & \cmark & \xmark & generated (87\%) + real (13\%) \\
3DreamBooth~\citep{ko2026threedreambooth} & 200K & multi-view (30) & \xmark & \xmark & static multi-view \\
\textbf{\method (ours)}                  & 4K    & multi-view (4) & \cmark & \cmark & generated (3D asset render) \\
\bottomrule
\end{tabular}
\end{table}

\subsection{Subject-Driven Video Generation}
\label{subsec:rw_subject}

Personalizing generative models to a user-specified subject originates in the image domain, through identifier-based personalization~\citep{ruiz2023dreambooth,gal2023textualinversion,kumari2023customdiffusion} and later encoder- or adapter-based conditioning that injects reference features without test-time optimization~\citep{ye2023ipadapter,tan2025ominicontrol}. These ideas carried over to video, where subject-to-video (S2V) methods keep a reference identity stable throughout a generated clip. Optimization- and adapter-based approaches~\citep{jiang2024videobooth,wei2024dreamvideo,he2024idanimator,yuan2025consisid} preserve identity over time, and reference-conditioned frameworks~\citep{liu2025phantom,deng2025magref,jiang2025vace,chen2025multisubject} extend this to multiple subjects and flexible prompts. These methods primarily encode the subject through 2D appearance features, leaving viewpoints not covered by the reference unconstrained.

Closest to our setting are recent attempts to move beyond single-view conditioning. MV-S2V~\citep{song2026mvs2v} feeds multiple reference views into a video diffusion transformer and uses a shifted positional-encoding scheme to distinguish different views of the same subject. The references provide appearance information from several viewpoints, but do not explicitly encode the 3D surface structure underlying those views. Its training videos for object-centric generation are produced by a camera-controllable video generator, so viewpoint variation is not explicitly tied to a shared 3D asset. 3DreamBooth~\citep{ko2026threedreambooth} uses static multi-view images for subject-specific adaptation, preserving the pretrained model's temporal behavior but requiring optimization for each new subject. In contrast, \method reconstructs geometry from the input views at inference and uses it directly as a conditioning signal, without subject-specific optimization. Requirements comparison across baselines are shown in Table~\ref{tab:requirements}.

%% file: sec/3_method.tex
\section{Method}
\label{sec:method}

\begin{figure}[t]
    \centering
    \includegraphics[width=\linewidth]{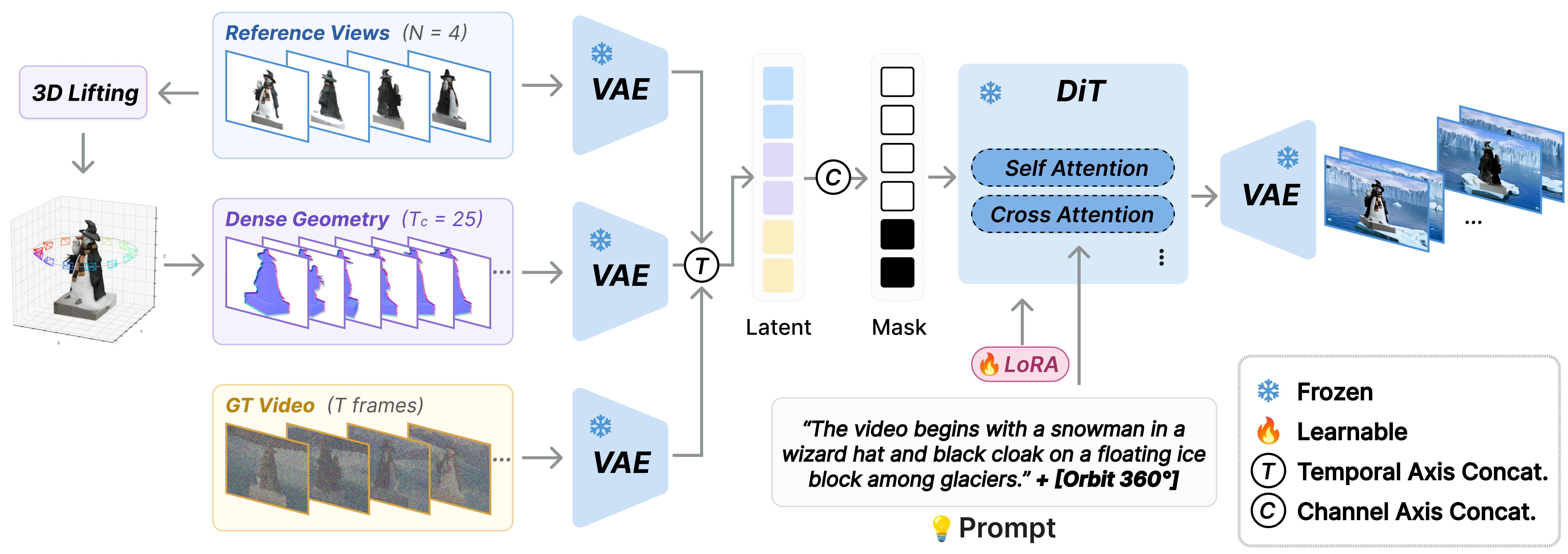}
    \caption{\textbf{Overview of \method.} From $N{=}4$ reference views, we reconstruct 3D geometry and render a surface-normal turntable. The reference images and normal condition are injected into the reference pathway of a LoRA-tuned image-to-video diffusion transformer. The model uses appearance information from the reference views and 3D structure from the normal condition to generate the target video without per-subject optimization.}
    \label{fig:method}
\end{figure}

Given a small set of reference views, we reconstruct the subject in 3D and render the reconstruction as a turntable of surface-normal maps. We provide these normal maps, together with the reference images, to a pretrained video diffusion model (Figure~\ref{fig:method}).

\subsection{Preliminaries and Problem Setup}
\label{subsec:method_pre}

\paragraph{Problem setup.}
A subject is given by a small set of reference views
$\mathcal{R}=\{\mathbf{I}_i\}_{i=1}^{N}$ ($N{=}4$ in our experiments), each
depicting the object from a different angle. Given a text prompt, we seek to
generate a video $\mathbf{V}=\{\mathbf{V}_t\}$ that follows the specified camera
trajectory and scene while preserving the subject's appearance and 3D structure
across viewpoints.
We further reconstruct an explicit 3D geometry from these
views and provide it to the generator, reducing its reliance on the backbone's
priors for viewpoints that are not directly observed.

\paragraph{Backbone.}
We build on a pretrained image-to-video (I2V) diffusion
transformer~\citep{wan2025} and adapt it with low-rank
tuning~\citep{hu2022lora}, keeping the base weights frozen. The model operates
in the latent space of a causal video VAE and is trained with a rectified-flow
objective. We denote the denoising transformer by $\mathcal{F}_\theta$, where
$\theta$ collects the trainable LoRA parameters.

\subsection{Automated 3D-Consistent Data Synthesis}
\label{subsec:method_data}

\begin{figure}[t]
    \centering
    \includegraphics[width=\linewidth]{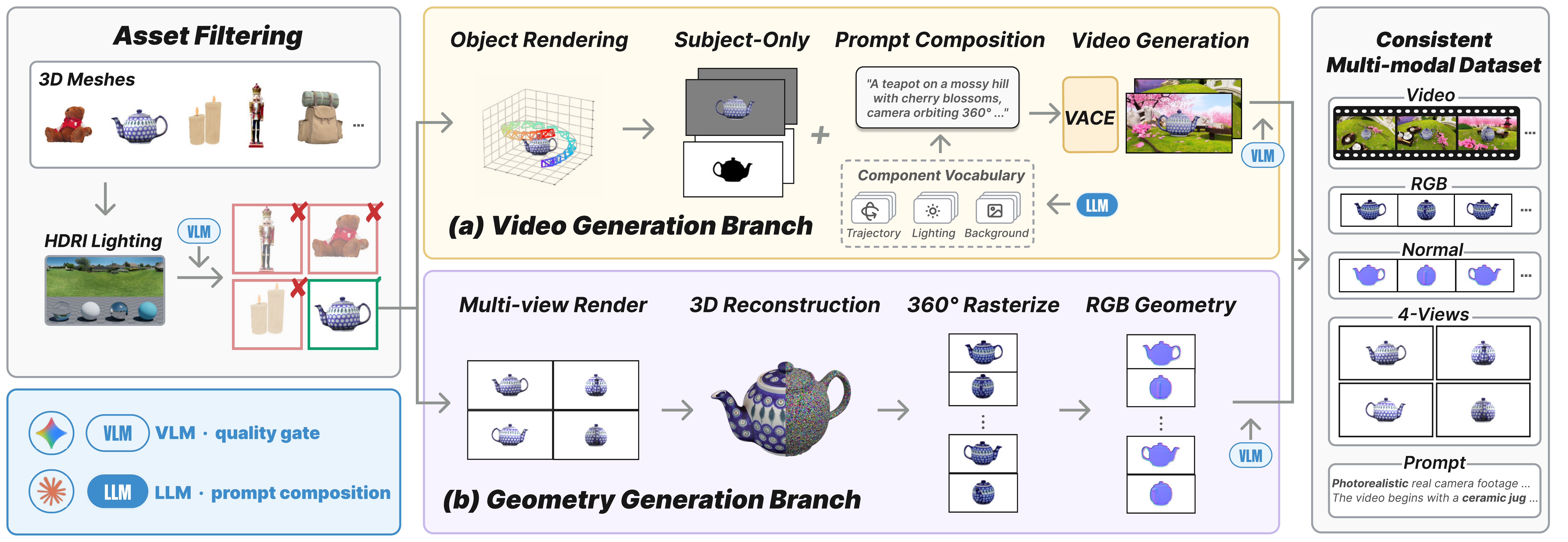}
    \caption{\textbf{Automated 3D-consistent data synthesis pipeline.} From a
curated 3D asset~\citep{deitke2023objaverse}, Blender renders the target orbit
and four reference views. (a) The target is composited into a prompt-driven
scene by video inpainting~\citep{jiang2025vace}; (b) the four views are
reconstructed in 3D~\citep{li2026mvsam3d} and rendered as the surface-normal
turntable used as the condition. Every stage runs without
human intervention.}
\label{fig:data}
\end{figure}

Our training data are designed to satisfy three requirements: subject identity preservation, geometrically consistent viewpoint changes, and explicit geometry for conditioning. A camera-controllable video generator does not guarantee these properties, because unseen surfaces are still synthesized by the generative prior. Real multi-view capture satisfies the first two but
leaves the third to an estimator. Rendering from a 3D asset satisfies all three
at once, because one mesh under known cameras produces every frame. However, direct rendering produces clean-background videos. We therefore add a separate scene-compositing stage to introduce realistic backgrounds corresponding to the camera motion.
The pipeline used to construct \textsc{Showcase-4K} has two branches: a
video-generation branch (Figure~\ref{fig:data}a) that produces the target video
and a geometry-generation branch (Figure~\ref{fig:data}b) that produces the
condition.

\begin{figure}[t]
    \centering
    \includegraphics[width=\linewidth]{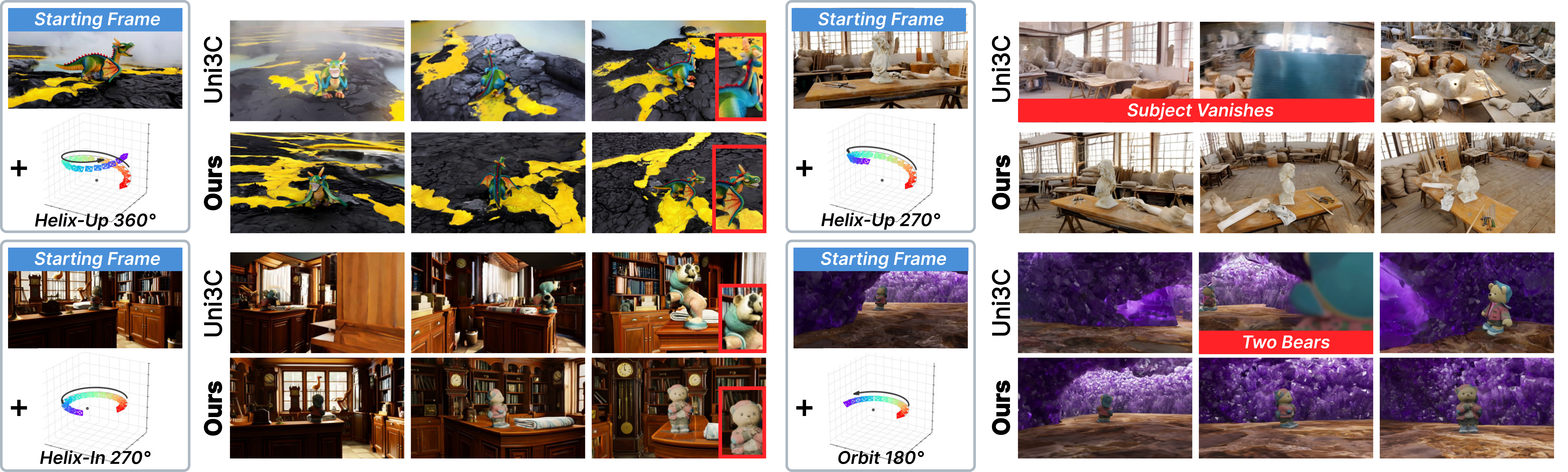}
    \caption{\textbf{Qualitative comparison of training data.}
    Uni3C~\citep{cao2025uni3c} is an image-to-video model, so every surface the
    camera path uncovers is generated. Ours renders the subject from a 3D asset
    and generates only the background.}
    \label{fig:datacompare}
\end{figure}

Figure~\ref{fig:datacompare} makes the difference concrete. Given one view and a camera path, a camera-controllable generator resolves every surface the first frame does not show from a prior, and within a single clip the subject can change shape, leave the scene, or appear twice at once.

\paragraph{3D-consistent reference and target views.}
We source 3D meshes from Objaverse~\citep{deitke2023objaverse} and, for
real-scanned subjects, from the Digital Twin Catalog~\citep{dong2025dtc} and
Google Scanned Objects~\citep{downs2022gso}. A heuristic pre-screen discards
degenerate meshes, and a vision-language model inspects four rendered views of
each remaining asset and keeps only those with valid geometry, plausible textures and
category consistency (rubrics in Appendix~\ref{app:rubric}). We use
four orbit-family motions (Orbit and Helix-Up/Down/In) with four rotation amounts each and eight additional
motions---arc, truck, dolly, and zoom variants---with nine framing transitions
each, yielding $88$ trajectory configurations in total
(Appendix~\ref{sec:traj-tags}). From the same mesh, Blender
renders an $81$-frame target video and $N{=}4$ reference views $\mathcal{R}$ at
$90^\circ$ yaw intervals, so references and target share one underlying shape
and appearance.

\paragraph{Subject-preserving compositing.}
We composite the subject into a prompt-driven scene to close the domain gap
from clean-background renders (Figure~\ref{fig:data}a). Subject masks
from the render's alpha channel drive a video inpainting
model~\citep{jiang2025vace} that synthesizes a coherent background around the
fixed subject, while keeping the rendered subject pixels unchanged. The scene prompt
combines a location from a language-model-written scene bank with the sample's
camera motion and lighting, and a
vision-language gate discards videos whose subject floats, deforms, or departs
from the references. The target is trimmed to $80$ frames to meet the video
VAE's temporal constraint (below).

\paragraph{Geometry conditions from reference views.}
Importantly, we reconstruct the geometry from the four reference views  (Figure~\ref{fig:data}b) rather than using the source mesh. This ensures that the conditioning signal available during training matches the signal available at inference.

We mask the subject on its white
background, predict aligned per-view pointmaps with a feed-forward multi-view
estimator~\citep{lin2025depthanything3},
and fuse the views into 3D Gaussians~\citep{kerbl20233dgs} with a multi-view
reconstruction module~\citep{li2026mvsam3d}. We render the Gaussians along a yaw
orbit at $15^\circ$ elevation at $T_c{=}25$ uniformly spaced poses and convert
each rendered depth map to surface normals. This
yields a normal turntable $\{\mathbf{N}_t\}_{t=1}^{T_c}$ that encodes the
surface visible from each viewpoint. A vision-language gate discards
reconstructions that depart from the reference or break apart.

\paragraph{Multimodal sample construction.}
A video-language model (Tarsier2-7B) captions the normal turntable and target video separately. Each training sample comprises a $25$-frame normal condition and an $80$-frame target, together with four reference images. The normal and target segments are encoded separately before being concatenated along the temporal axis in latent space, while each reference image is encoded independently. Videos are sampled at $16$~fps from the renders. The two segments are paired with explicit \texttt{[CONDITION]} and \texttt{[VIDEO]} markers describing the normal turntable and target video, respectively.

\subsection{Geometry-Aware Conditioning}
\label{subsec:method_cond}

\paragraph{Conditioning inputs.}
The conditioning sequence holds the $N{=}4$ reference images followed by the
$T_c{=}25$-frame normal turntable, and the target segment follows as the frames
to be produced. The target camera motion is specified by the text prompt, while the normal turntable provides a trajectory-independent geometry cue.

We encode the two conditions differently because they have different structures. Each reference image is encoded independently, while the normal maps form a temporal sequence and are encoded using the video VAE.

\paragraph{Geometry-aware in-context conditioning.}
The sequence concatenates the reference, normal and target
latents along the temporal axis and the transformer attends over it,
\begin{equation}
\mathbf{z}=\big[\,\mathbf{z}^{\mathcal{R}};\;\mathbf{z}^{\mathcal{N}};\;\mathbf{z}^{\mathcal{V}}\,\big]
\in\mathbb{R}^{L\times C\times h\times w},
\qquad L=N+T_n+T_v.
\end{equation}
Encoding the $25$-frame normal condition and the $80$-frame target separately
yields $T_n{=}7$ and $T_v{=}20$ latent frames, respectively. Together with the
$N{=}4$ independently encoded reference images, the resulting sequence has
length $L=N+T_n+T_v=31$. A binary mask $\mathbf{m}\in\{0,1\}^{L}$ marks the first $N+T_n$ positions as given. In element-wise operations, $\mathbf{m}$ is broadcast over the channel and spatial dimensions. For channel-wise concatenation, we expand it into $\mathbf{M}\in\{0,1\}^{L\times4\times h\times w}$. The transformer input stacks three streams along the channel axis:
\begin{align}
\widetilde{\mathbf{z}}_\tau
&=\mathbf{m}\odot\mathbf{z}
+(1-\mathbf{m})\odot
\big(\tau\mathbf{z}+(1-\tau)\boldsymbol{\epsilon}\big),
\quad \boldsymbol{\epsilon}\sim\mathcal{N}(\mathbf{0},\mathbf{I}),\\[-2pt]
\mathbf{x}_\tau
&=\operatorname{Concat}_{\mathrm{channel}}\!\left(\widetilde{\mathbf{z}}_\tau,\mathbf{M},\mathbf{m}\odot\mathbf{z}\right).
\end{align}
Thus, the reference and normal latents remain clean, while only the target-video
latents are interpolated with noise. The other two streams provide the mask and
the clean conditioning latents zeroed over the target span. The mask states
which positions were supplied, and the rectified-flow loss is applied where
$m_\ell=0$. The four channels of $\mathbf{M}$ follow the VAE temporal grouping. With $C=16$ latent channels, the three input streams therefore contribute $16+4+16=36$ channels.

No frame-wise correspondence between the normal and target sequences is imposed.
This is why the turntable enters as its own span rather than as extra channels on
the target latents. A channel-wise condition has to be read frame by frame, which
would force an alignment between the $25$ normals and the $80$ target frames that
does not exist, since the turntable is a fixed yaw sweep and the target follows
whichever trajectory the prompt names.

\subsection{Training and Inference}
\label{subsec:method_train}

\paragraph{Training.}
We optimize only the LoRA parameters $\theta$ with the backbone's rectified-flow
objective. Let $\mathbf{c}$ denote the text condition and
$\mathcal{F}_\theta^{\mathcal{V}}$ the target-video segment of the transformer
output. Writing $\widehat{\mathbf{u}}_\theta^{\mathcal{V}}
=\mathcal{F}_\theta^{\mathcal{V}}(\mathbf{x}_\tau,\tau\mid\mathbf{c})$, the
target velocity under the backbone's flow convention is
$\mathbf{z}^{\mathcal{V}}-\boldsymbol{\epsilon}^{\mathcal{V}}$. We minimize
\begin{equation}
\mathcal{L}
=\mathbb{E}\!\left[
\frac{1}{D_v}\left\lVert
\widehat{\mathbf{u}}_\theta^{\mathcal{V}}
-\big(\mathbf{z}^{\mathcal{V}}-\boldsymbol{\epsilon}^{\mathcal{V}}\big)
\right\rVert_2^2
\right],
\end{equation}
where $D_v$ is the number of scalar elements in $\mathbf{z}^{\mathcal{V}}$, and the expectation is over $\mathbf{z}$, $\boldsymbol{\epsilon}$, $\tau$, and
$\mathbf{c}$. The squared error is averaged over target-video latent elements only;
$\mathbf{z}^{\mathcal{R}}$ and $\mathbf{z}^{\mathcal{N}}$ remain clean context
and do not contribute to the loss.

\paragraph{Inference.}
At test time we run the same reconstruction and rendering steps on a new
subject's reference views (reconstruct the subject, render the yaw orbit, and
derive the surface-normal turntable), then condition the generator on the
references and normals to produce the target latents. The $7$ normal-condition
latents and $20$ generated target latents are concatenated and jointly decoded.
Under the causal VAE's temporal compression factor of $4$, the resulting
$27$-frame latent sequence decodes into $1+4(27-1)=105$ frames: $25$
normal-condition frames followed by $80$ target frames.

%% file: sec/4_experiment.tex
\section{Experiment}
\label{sec:experiment}

\subsection{Experimental Setup}
\label{subsec:setup}

\noindent\textbf{Datasets and baselines.}
We evaluate on \textsc{ShowcaseBench}, a held-out benchmark of $50$ subjects
spanning four $360^\circ$ camera trajectories and sharing no 3D asset with
\textsc{Showcase-4K}, and on NAVI~\citep{jampani2023navi} with $35$ subjects.
Training and inference settings are given in Appendix~\ref{app:hyperparams}.
Phantom~\citep{liu2025phantom} and MAGREF~\citep{deng2025magref} are single-view
S2V models, and we run them in the multi-reference mode. We refer to these
variants as Phantom-MV and MAGREF-MV, where the four reference images are treated
as views of the same subject. Appendix~\ref{app:baseline-ext} also runs both in
their original single-view mode, where the four-view extension is the stronger
setting on almost every metric, so the multi-view numbers are the ones we report. We further compare against recent methods,
MV-S2V~\citep{song2026mvs2v} and 3DreamBooth~\citep{ko2026threedreambooth}.
All methods in our main comparison use the same four reference images.
For 3DreamBooth, these images are used for both subject-specific optimization and inference.
Prompt details are provided in Appendices~\ref{app:rubric_caption}--\ref{app:rubric_prompts}.

\noindent\textbf{Metrics.} We sample $16$ frames per video and obtain subject masks for the reference images and sampled video frames using GroundingDINO~\citep{liu2024groundingdino} and SAM~\citep{kirillov2023sam}.
Every identity and 3D metric is computed on those masked subjects, so background
content does not enter them, while text alignment and video quality use full frames.

\emph{Identity.} DINO and CLIP similarity measure how well the generated orbit
covers the supplied reference views. For each reference view we take the maximum similarity over generated frames and
average across references. Features come from
DINOv2-L~\citep{oquab2024dinov2} and CLIP ViT-L/14~\citep{radford2021clip}.

\emph{3D consistency.} MEt3R~\citep{asim2025met3r} scores multi-view consistency
between reference and generated views, taking the minimum dissimilarity per
reference. We also lift the generated orbit to a point cloud with
$\pi^3$~\citep{wang2025pi3} and compare it with the dataset's ground-truth
geometry. For \textsc{ShowcaseBench}, the ground-truth point cloud is
back-projected from source-mesh depth renders; for NAVI, it is sampled from the
3D scan mesh. We report GT-CD as the symmetric mean nearest-neighbor Euclidean distance between the predicted and ground-truth point clouds, following independent centering, mean-radius normalization, and rigid alignment. We additionally report \emph{RotCoverage} as an alignment-dependent diagnostic of viewpoint coverage, computed as the mean distance from each reference viewpoint to the nearest estimated generated viewpoint after pose alignment. Appendix~\ref{app:hyperparams} gives the evaluation details.

\emph{Text alignment.} We compute ViCLIP~\citep{wang2024internvid} cosine
similarity between the video and the prompt using eight sampled frames and the
clause describing the target video.

\emph{Video quality.} We report VBench~\citep{huang2024vbench} imaging quality,
aesthetic quality and dynamic degree for completeness.

\subsection{Qualitative and Quantitative Comparison}
\label{subsec:qual}

\begin{figure}[t]
    \centering
    \includegraphics[width=\linewidth]{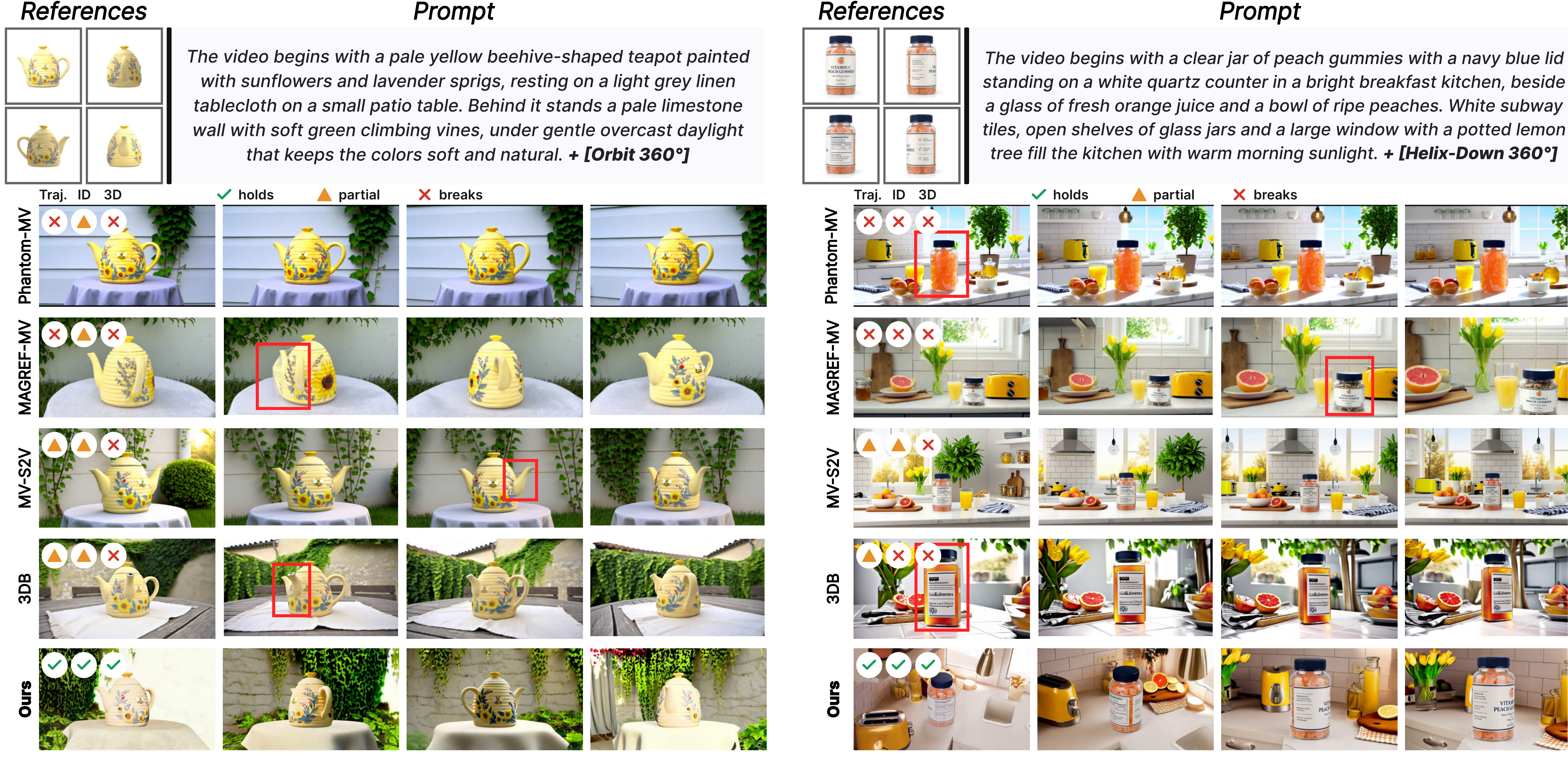}
    \caption{\textbf{Qualitative comparison with baselines.} 3DB denotes 3DreamBooth.
    Given a teapot and a jar of peach gummies, we compare \method against four
    baselines. Red boxes mark where identity or 3D structure breaks down.}
    \label{fig:comparison}
\end{figure}

Figure~\ref{fig:comparison} shows how each baseline breaks down. Each method
carries three marks for whether the camera follows the trajectory the prompt
names (Traj.), whether the subject keeps its identity (ID) and whether it stays
3D-consistent (3D). Phantom-MV and MAGREF-MV barely orbit the subject, and the few faces they do
reveal come apart. MV-S2V likewise stops short of completing the rotation, and
within one clip it grows a second spout on the teapot and snaps back to an
earlier frame, so the sequence does not read as one object seen from moving
viewpoints. 3DreamBooth converges on neither the subject identity nor its
geometry, leaving both off-target.

\begin{table}[t]
\centering
\small
\setlength{\tabcolsep}{4pt}
\renewcommand{\arraystretch}{1.15}
\caption{\textbf{Main quantitative comparison.} Results on \textsc{ShowcaseBench}
with $50$ subjects and on NAVI~\citep{jampani2023navi} with $35$
subjects. We evaluate identity preservation, 3D consistency, text alignment and
video quality through nine metrics.
Best results are in \textbf{bold} and second best are \underline{underlined}.}
\label{tab:main}
\resizebox{\linewidth}{!}{%
\begin{tabular}{ll cc ccc c ccc}
\toprule
 & & \multicolumn{2}{c}{\textbf{Identity}}
   & \multicolumn{3}{c}{\textbf{3D consistency}}
   & \multicolumn{1}{c}{\textbf{Text alignment}}
   & \multicolumn{3}{c}{\textbf{Video quality}} \\
\cmidrule(lr){3-4}\cmidrule(lr){5-7}\cmidrule(lr){8-8}\cmidrule(lr){9-11}
Set & Method
 & DINO$\uparrow$ & CLIP$\uparrow$
 & MEt3R$\downarrow$ & GT-CD$\downarrow$ & RotCov$\downarrow$
 & ViCLIP$\uparrow$
 & Imaging$\uparrow$ & Aesthetic$\uparrow$ & Dynamic$\uparrow$ \\
\midrule
\multirow{5}{*}{\rotatebox{90}{Held-out}}
 & Phantom-MV                                   & 0.621 & 0.871 & 0.444 & 0.2843 & 0.906 & 0.236 & 72.52 & 0.595 & 0.460 \\
 & MAGREF-MV                                    & 0.616 & 0.876 & 0.409 & 0.1789 & 1.177 & 0.242 & \underline{73.88} & \textbf{0.614} & 0.760 \\
 & MV-S2V                                       & 0.688 & 0.896 & 0.351 & 0.2306 & 0.787 & \underline{0.244} & 73.21 & 0.590 & \underline{0.960} \\
 & 3DreamBooth                                  & \underline{0.707} & \underline{0.907} & \underline{0.325} & \underline{0.1717} & \underline{0.765} & 0.238 & 70.67 & \underline{0.598} & \textbf{1.000} \\
 & \textbf{\method (ours)}                      & \textbf{0.799} & \textbf{0.925} & \textbf{0.190} & \textbf{0.0824} & \textbf{0.665} & \textbf{0.259} & \textbf{74.29} & 0.574 & \textbf{1.000} \\
\midrule
\multirow{5}{*}{\rotatebox{90}{NAVI}}
 & Phantom-MV                                   & 0.690 & 0.882 & 0.462 & 0.2014 & 1.458 & \underline{0.255} & 71.17 & \underline{0.572} & 0.029 \\
 & MAGREF-MV                                    & 0.665 & 0.892 & 0.457 & 0.2145 & 1.284 & 0.251 & 69.81 & \textbf{0.574} & \underline{0.343} \\
 & MV-S2V                                       & \underline{0.770} & \underline{0.912} & \underline{0.350} & \underline{0.1816} & \underline{1.163} & 0.250 & \underline{71.73} & 0.565 & \textbf{1.000} \\
 & 3DreamBooth                                  & 0.741 & \underline{0.912} & 0.386 & 0.2272 & 1.179 & 0.244 & 70.34 & 0.559 & \textbf{1.000} \\
 & \textbf{\method (ours)}                      & \textbf{0.828} & \textbf{0.918} & \textbf{0.249} & \textbf{0.0975} & \textbf{1.074} & \textbf{0.260} & \textbf{74.58} & 0.551 & \textbf{1.000} \\
\bottomrule
\end{tabular}}
\par\smallskip\raggedright\footnotesize

\end{table}

Table~\ref{tab:main} compares both benchmarks, and the weaknesses it exposes follow the qualitative ones.
No baseline is strong on every axis, with Phantom-MV and MAGREF-MV trailing on identity and camera travel and the lead between MV-S2V and 3DreamBooth changing across the two benchmarks. Overall, \method achieves the best DINO, CLIP, MEt3R, and GT-CD scores on both benchmarks while keeping text alignment and video quality competitive.

\subsection{Ablation}
\label{subsec:ablation}

\begin{figure}[t]
    \centering
    \includegraphics[width=\linewidth]{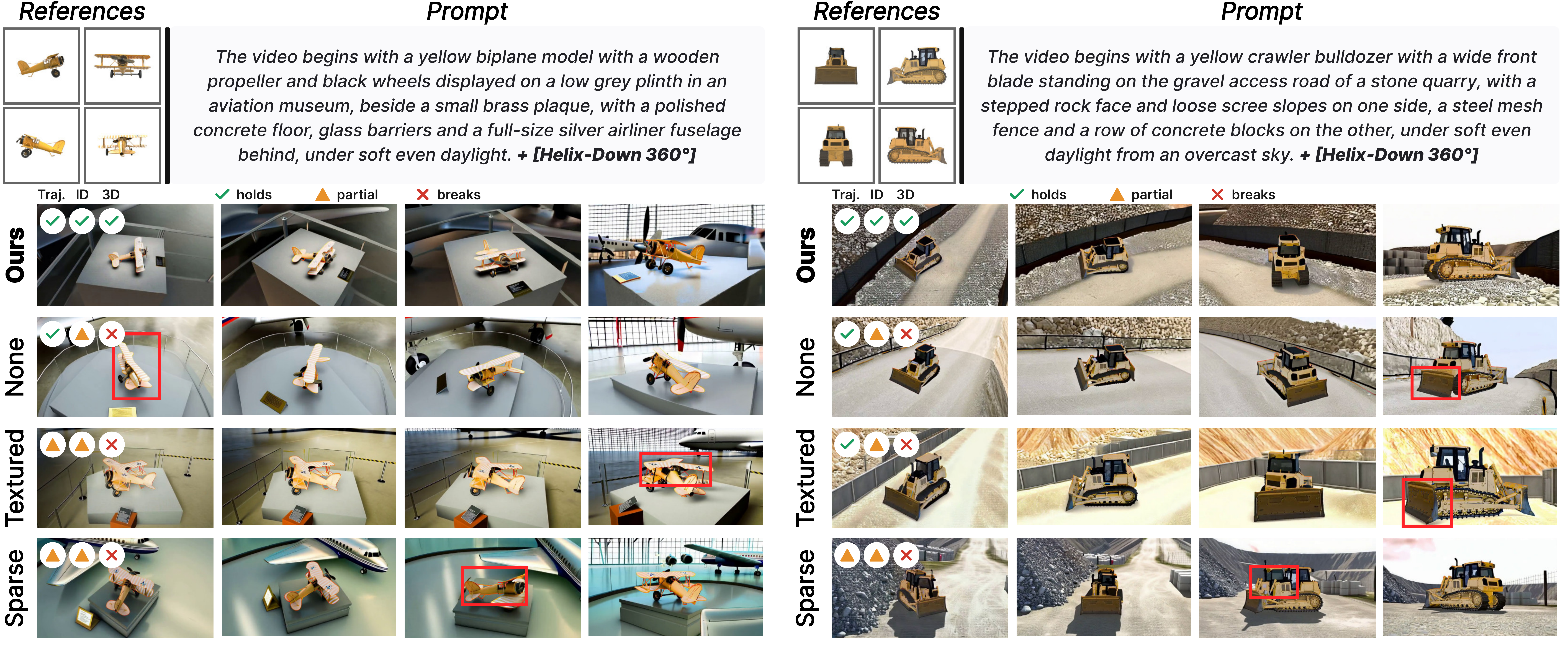}
    \caption{\textbf{Conditioning ablation.} Given a biplane and a bulldozer,
    we compare the three conditioning variants against ours. Marked regions
    highlight where identity or 3D structure breaks down.}
    \label{fig:ablation}
\end{figure}

\begin{table}[t]
\centering
\small
\setlength{\tabcolsep}{4pt}
\renewcommand{\arraystretch}{1.15}
\caption{\textbf{Geometry-condition ablation.} None keeps only the
reference views, Textured replaces the normals with a colored render of the
same reconstruction, and Sparse thins the turntable from $25$ to $9$ views. Reference count is varied
separately in Table~\ref{tab:st_nview}. Best results are in \textbf{bold} and
second best are \underline{underlined}.}
\label{tab:ablation}
\resizebox{\linewidth}{!}{%
\begin{tabular}{ll cc ccc c cc}
\toprule
 & & \multicolumn{2}{c}{\textbf{Identity}}
   & \multicolumn{3}{c}{\textbf{3D consistency}}
   & \multicolumn{1}{c}{\textbf{Text}}
   & \multicolumn{2}{c}{\textbf{Video quality}} \\
\cmidrule(lr){3-4}\cmidrule(lr){5-7}\cmidrule(lr){8-8}\cmidrule(lr){9-10}
Set & Geometry condition
 & DINO$\uparrow$ & CLIP$\uparrow$
 & MEt3R$\downarrow$ & GT-CD$\downarrow$ & RotCov$\downarrow$
 & ViCLIP$\uparrow$
 & Imaging$\uparrow$ & Aesthetic$\uparrow$ \\
\midrule
\multirow{4}{*}{\rotatebox{90}{Held-out}}
 & \textbf{Normals (ours)}   & \textbf{0.799} & \textbf{0.925} & \textbf{0.190} & \textbf{0.0824} & \textbf{0.665} & \textbf{0.259} & \textbf{74.29} & \textbf{0.574} \\
 & None                      & 0.752 & 0.903 & 0.265 & 0.2092 & 0.727 & 0.248 & 69.85 & 0.547 \\
 & Textured                  & \underline{0.755} & 0.910 & 0.270 & \underline{0.1698} & 0.747 & 0.236 & \underline{73.41} & \underline{0.564} \\
 & Sparse                    & 0.752 & \underline{0.912} & \underline{0.263} & 0.1777 & \underline{0.698} & \underline{0.250} & 71.84 & 0.552 \\
\midrule
\multirow{4}{*}{\rotatebox{90}{NAVI}}
 & \textbf{Normals (ours)}   & \textbf{0.828} & \textbf{0.918} & \textbf{0.249} & \textbf{0.0975} & \textbf{1.074} & \textbf{0.260} & \textbf{74.58} & \textbf{0.551} \\
 & None                      & \underline{0.793} & 0.904 & 0.315 & 0.1987 & 1.107 & 0.238 & 67.67 & 0.532 \\
 & Textured                  & 0.780 & 0.907 & 0.323 & 0.1842 & \underline{1.089} & \underline{0.240} & \underline{71.47} & \underline{0.544} \\
 & Sparse                    & 0.772 & \underline{0.909} & \underline{0.288} & \underline{0.1769} & 1.090 & \underline{0.240} & 67.72 & 0.531 \\
\bottomrule
\end{tabular}}
\end{table}

Table~\ref{tab:ablation} ablates the conditioning signal in three ways, holding
the backbone and the training data fixed: (i) removing geometry conditioning
entirely and keeping only the reference views, (ii) replacing the surface-normal
turntable with a textured render of the same reconstruction, and (iii) thinning
the dense turntable from $25$ to $9$ frames. Each variant is trained
independently from the same initialization and evaluated at $33{,}000$ steps.

Without geometry conditioning the model keeps the reference appearance but is
told nothing about how the subject is shaped away from those views, so identity preservation, MEt3R, and GT-CD all deteriorate. The auxiliary RotCoverage score also increases. The textured variant carries the same geometry but paints it
with reconstructed colour, which competes with the reference images for
appearance: identity barely moves relative to dropping the condition, and text
alignment falls furthest of the three on \textsc{ShowcaseBench}. Thinning the
turntable preserves the modality but not the coverage, and its cost depends on
how reliable the reconstruction is. On \textsc{ShowcaseBench}, Textured gives the
lowest GT-CD among the three ablations, while Sparse gives the lowest auxiliary RotCoverage score. On NAVI, removing geometric conditioning yields the highest GT-CD. Figure~\ref{fig:ablation}
shows the same comparison qualitatively.

%% file: sec/5_conclusion.tex
\section{Conclusion}

We present \method, which pairs an automated pipeline that renders omnidirectional
videos from 3D assets with surface-normal turntable conditioning reconstructed from
the reference views, requiring no temporal alignment with generated frames. \method
leads single- and multi-view baselines on subject identity and cross-view 3D
consistency on both \textsc{ShowcaseBench} and NAVI. Two limitations remain: the
background sometimes looks synthetic, and subjects that are not rigid across the four
references are not handled (Figure~\ref{fig:sf_failure}, Appendix~\ref{app:failure}).
We leave both to future work. We will release the data, benchmark, and code.

%% file: sec/X_suppl.tex
\clearpage
\section*{Supplementary Material}
\addcontentsline{toc}{section}{Supplementary Material}

\section{Additional Qualitative Results}
\label{app:qual}

\begin{figure}[h]
    \centering
    \includegraphics[width=\linewidth]{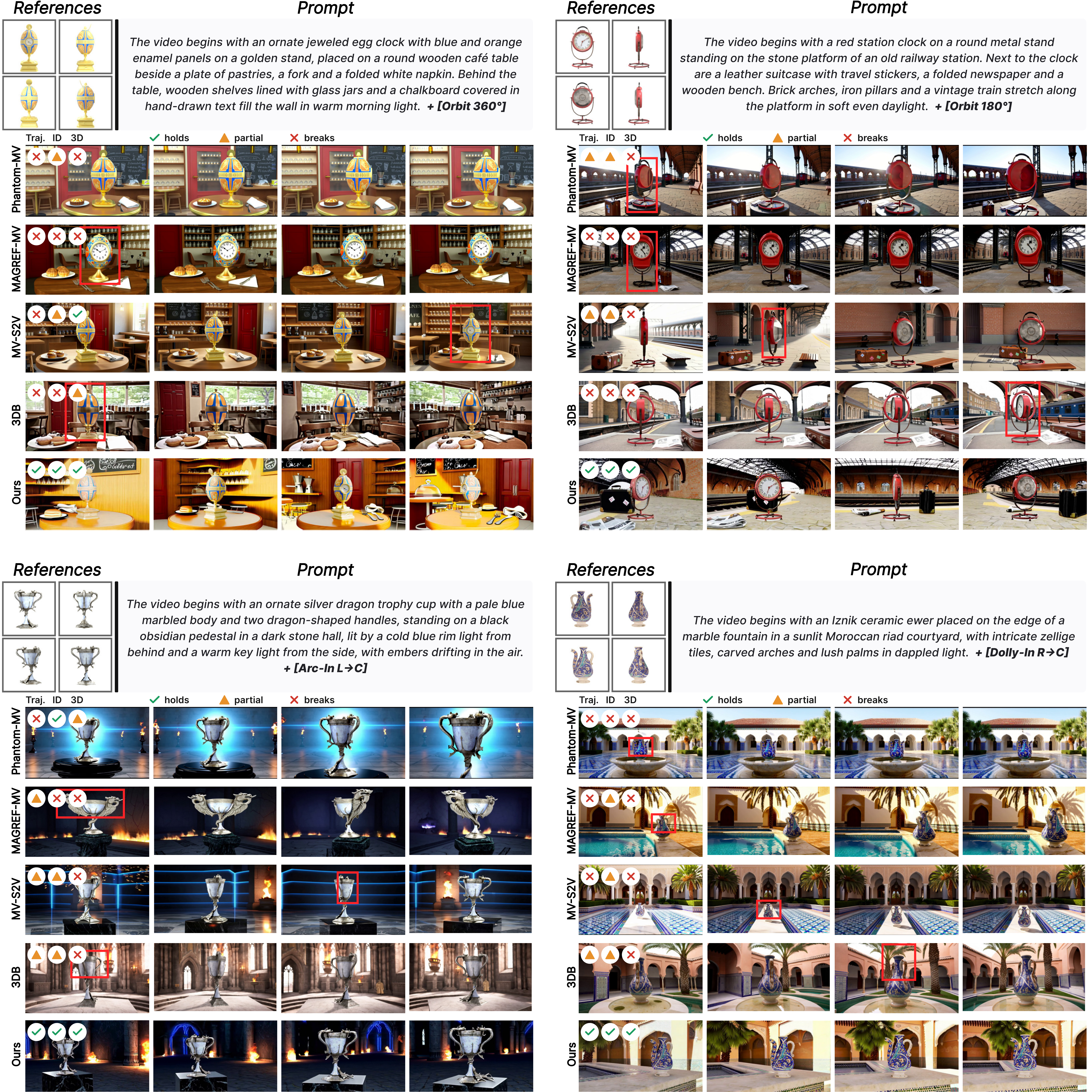}
    \caption{\textbf{Additional qualitative comparison.} Four further subjects
    under diverse camera trajectories, with the same method ordering as
    Figure~\ref{fig:comparison}. Each block pairs the four reference views and
    the prompt, whose \texttt{+\,[tag]} suffix names the trajectory
    (Appendix~\ref{sec:traj-tags}), with generated frames from every method.}
    \label{fig:sf_quali2}
\end{figure}

Figure~\ref{fig:sf_quali2} extends the comparison of Figure~\ref{fig:comparison} to
four more subjects and a different trajectory each, with the methods in the same
order. Consistent with the quantitative results, \method completes every commanded
trajectory while holding the same object throughout. In contrast, Phantom-MV and MAGREF-MV
rotate only slightly, so their later frames fall back on views the references already
provide, MV-S2V rotates further but reconstructs the subject as it travels, and
3DreamBooth retains the silhouette while losing the surface detail that identifies
the particular object.

\begin{figure}[h]
    \centering
    \includegraphics[width=\linewidth]{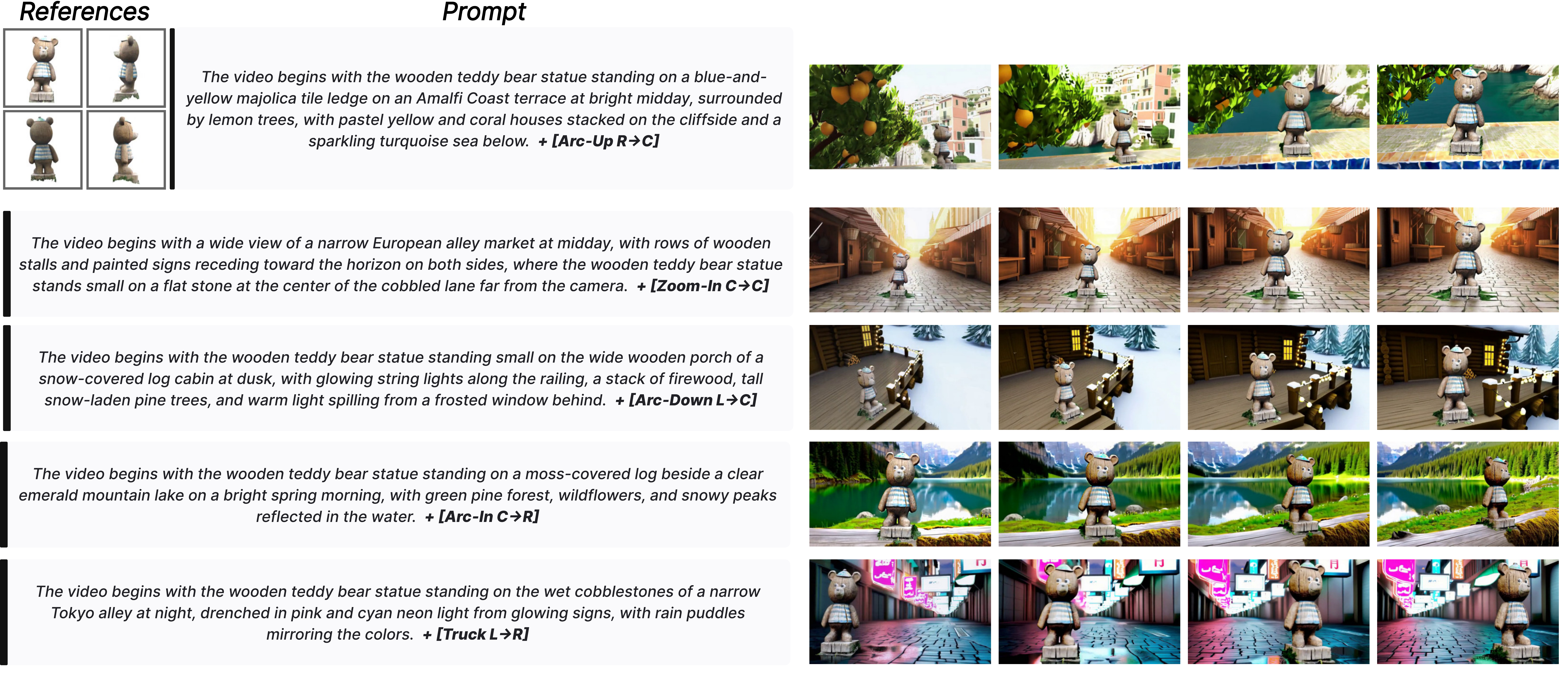}
    \caption{\textbf{Trajectory and scene control.}
    One subject driven by several camera configurations, including arc, zoom-in
    and truck, with the background also varied by the prompt for each
    trajectory.}
    \label{fig:sf_beyond}
\end{figure}

\begin{figure}[h]
    \centering
    \includegraphics[width=\linewidth]{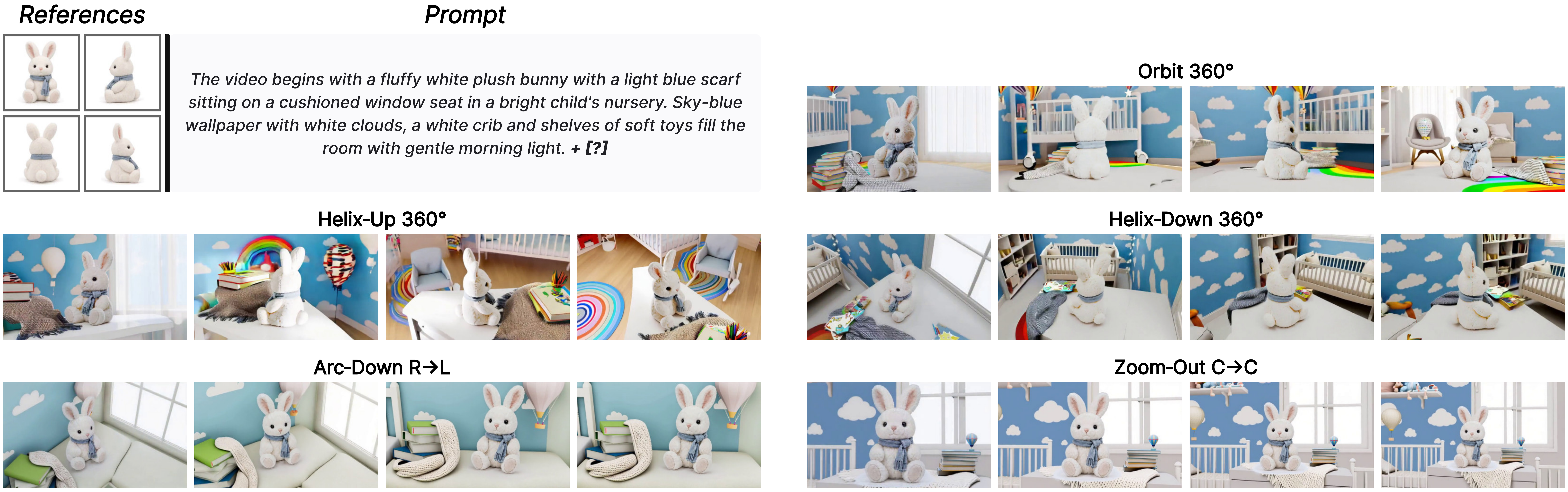}
    \caption{\textbf{Trajectory control in a fixed scene.} The same subject and
    background under different camera configurations, from full orbits to an
    arc and a zoom-out, where only the camera trajectory changes between rows.
    Inference only.}
    \label{fig:sf_beyond_bg}
\end{figure}

Figures~\ref{fig:sf_beyond} and~\ref{fig:sf_beyond_bg} examine how far the camera
follows the prompt once the subject is fixed. The corpus spans the full motion
vocabulary of Table~\ref{tab:traj-tags}, so the question is not whether a motion was
seen but whether naming it is enough to obtain it. In Figure~\ref{fig:sf_beyond} one
subject is driven by an arc, a zoom-in and a truck while the prompt also changes the
scene, so the path and the surroundings vary together.
Figure~\ref{fig:sf_beyond_bg} isolates the camera by holding the subject and the
background fixed and changing only the trajectory. The subject keeps the same
identity and shape throughout while the path it is viewed along changes.

\begin{figure}[h]
    \centering
    \includegraphics[width=\linewidth]{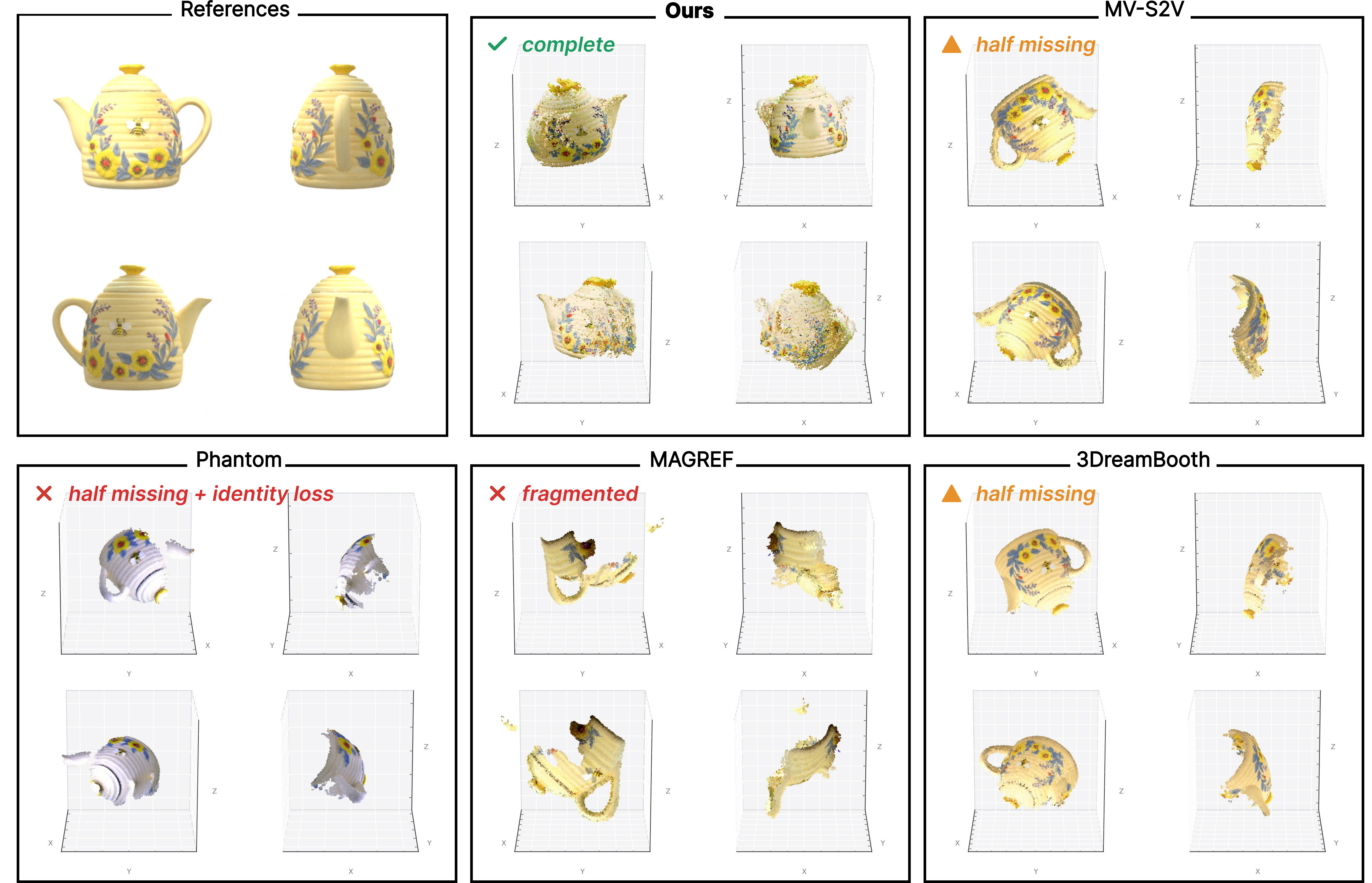}
    \caption{\textbf{3D reconstruction from generated video.} The subject is
    segmented out of each method's generated orbit and lifted to a coloured point
    cloud, the same reconstruction evaluated by GT-CD, shown here
    from four viewpoints so the geometric differences are visible directly.
    Baselines recover a thin shell of the surfaces their orbit actually sweeps,
    while ours closes the volume.}
    \label{fig:sf_3drecon}
\end{figure}

Figure~\ref{fig:sf_3drecon} lifts each generated orbit into a point cloud with the same
reconstruction used for GT-CD and shows it from four viewpoints, so the
difference the metric reports is visible as shape. The baselines produce a shell that
covers only the arc their camera actually swept and opens up everywhere else, while
\method closes the volume, which is what a metric computed against ground-truth
geometry is measuring.

\section{User Study}
\label{app:userstudy}

\begin{figure}[h]
    \centering
    \includegraphics[width=\linewidth]{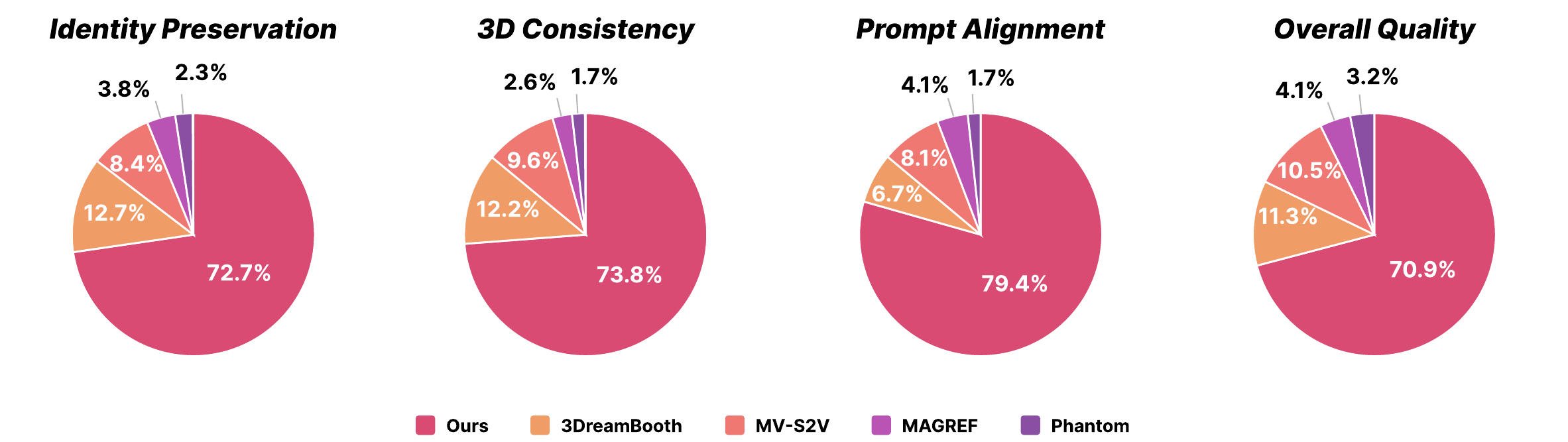}
    \caption{\textbf{User study results.} \method is chosen most often on all
    four criteria, ahead of every baseline by a wide margin.}
    \label{fig:userstudy}
\end{figure}

To check that the automatic metrics agree with what people actually see, we ran
a preference study involving $24$ videos that are not seen during training. A
total of $44$ participants were shown the reference views, the prompt and the
five results side by side, our method and the four baselines, and selected one
video for each of the following criteria.

\begin{itemize}[leftmargin=1.4em,itemsep=1pt,topsep=2pt]
\item \textbf{Identity Preservation.} The video that best preserves the identity
of the subject across viewpoints.
\item \textbf{3D Consistency.} The video that keeps a consistent 3D shape and
structure while the subject turns.
\item \textbf{Prompt Alignment.} The video that best follows the prompt.
\item \textbf{Overall Quality.} The video with the best overall quality.
\end{itemize}

Figure~\ref{fig:userstudy} reports the share of selections. \method is preferred
on every criterion, taking between $70.9\%$ and $79.4\%$ of the votes while no
baseline passes $12.7\%$. The margin is largest on prompt alignment,
which is what the clips show most plainly, since \method is the only method
whose camera completes the commanded orbit rather than turning a few degrees or staying still.

\section{Implementation Details}
\label{app:hyperparams}

We fine-tune Wan2.1 I2V-14B-480P through its Diffusers implementation. Training
applies LoRA to the DiT with rank $128$ and $\alpha=128$, and every other weight
stays frozen. Each training sample contains a $25$-frame normal-condition segment and an $80$-frame target segment at $480\times832$ resolution. These segments are encoded separately into $7$ and $20$ latent frames, respectively, and concatenated along the temporal axis to form a $27$-frame latent sequence. The four in-context reference images are encoded independently. We train for $33{,}000$
steps at a batch size of $2$ per device over four NVIDIA B200 GPUs with no
gradient accumulation, which takes about $50$ hours. The optimizer is AdamW with $\beta=(0.9,\,0.95)$ and
$\epsilon=10^{-8}$, weight decay $10^{-4}$ and gradient clipping at $1.0$. The
learning rate is $2\times10^{-5}$, held constant after $100$ warmup steps.
Training runs in bfloat16 under seed $42$, and every variant reported in this
paper uses the same $33{,}000$-step checkpoint.

At inference, we take $50$ sampling steps with three-way classifier-free
guidance, weighting the reference stream at $s_\text{ref}=5.0$ and the text
stream at $s_\text{text}=5.0$. The concatenated $27$-frame latent sequence is jointly decoded into $1+4(27-1)=105$ frames, comprising $25$ normal-condition frames followed by $80$ target frames at $480\times832$ resolution. Clips are written at $30$~fps.

For GT-CD, the ground-truth point cloud is back-projected from source-mesh depth renders for \textsc{ShowcaseBench} and sampled from the 3D scan mesh for NAVI. We independently center the predicted and ground-truth clouds at their respective centroids and normalize each by its mean Euclidean radius. We then align the normalized prediction to the ground truth using rigid ICP with scale fixed. The evaluation removes the outermost $3\%$ of points by radius and subsamples each cloud to at most $20{,}000$ points. For the resulting point clouds $P$ and $Q$, we compute
\[
d(P,Q)=\frac{1}{|P|}\sum_{p\in P}\min_{q\in Q}\lVert p-q\rVert_2,\qquad
\mathrm{GT\mbox{-}CD}=\tfrac{1}{2}\big(d(P,Q)+d(Q,P)\big).
\]
The directed terms measure prediction-to-GT accuracy and GT-to-prediction completeness using Euclidean rather than squared distances.

RotCoverage is an auxiliary, alignment-dependent diagnostic of viewpoint coverage. After similarity alignment, the generated and reference camera poses are normalized by the RMS radius of the reference camera centers. Each pose is represented by five frustum vertices: the apex and four far-plane corners at depth $0.5$ and half-size $0.25$. Pose distance is the mean Euclidean distance between corresponding frustum vertices. For each reference pose, we take the minimum distance over generated poses and average across references, yielding a dimensionless score.

\section{Reference-count Ablation}
\label{app:nview}

\begin{table}[h]
\centering
\small
\setlength{\tabcolsep}{4pt}
\renewcommand{\arraystretch}{1.15}
\caption{\textbf{Reference-count flexibility.} Although \method assumes four
reference views, the model generalizes to fewer, while the full four-view setting
performs best overall. The condition slots are filled by cycling the available
views, and the surface-normal turntable is re-lifted from exactly those $n$
views, so nothing about the missing views leaks in. Every row uses the same
checkpoint and the same evaluation protocol as Table~\ref{tab:main}, and
$n\!=\!4$ repeats the \method row of that table.}
\label{tab:st_nview}
\resizebox{\linewidth}{!}{%
\begin{tabular}{l cccccc cccccc}
\toprule
 & \multicolumn{6}{c}{\textsc{ShowcaseBench}} & \multicolumn{6}{c}{NAVI} \\
\cmidrule(lr){2-7}\cmidrule(lr){8-13}
\#refs
 & DINO$\uparrow$ & CLIP$\uparrow$ & MEt3R$\downarrow$ & GT-CD$\downarrow$ & RotCov$\downarrow$ & ViCLIP$\uparrow$
 & DINO$\uparrow$ & CLIP$\uparrow$ & MEt3R$\downarrow$ & GT-CD$\downarrow$ & RotCov$\downarrow$ & ViCLIP$\uparrow$ \\
\midrule
$n=1$        & 0.702 & 0.902 & 0.301 & 0.1704 & 0.691 & \underline{0.258}
             & 0.811 & 0.901 & 0.307 & \underline{0.1875} & 1.091 & 0.245 \\
$n=2$        & 0.768 & 0.918 & 0.241 & 0.1658 & 0.685 & \underline{0.258}
             & 0.814 & \underline{0.913} & 0.262 & 0.1915 & \underline{1.075} & \underline{0.249} \\
$n=3$        & \underline{0.783} & \underline{0.921} & \underline{0.206} & \underline{0.1480} & \underline{0.670} & 0.256
             & \underline{0.823} & 0.911 & \underline{0.255} & 0.2050 & \textbf{1.074} & \underline{0.249} \\
$n=4$ (full) & \textbf{0.799} & \textbf{0.925} & \textbf{0.190} & \textbf{0.0824} & \textbf{0.665} & \textbf{0.259}
             & \textbf{0.828} & \textbf{0.918} & \textbf{0.249} & \textbf{0.0975} & \textbf{1.074} & \textbf{0.260} \\
\bottomrule
\end{tabular}}
\end{table}

The metrics separate by how much each one depends on unseen geometry. The full
four-view setting gives the lowest GT-CD on both \textsc{ShowcaseBench} and NAVI,
while the NAVI identity scores stay nearly flat across reference counts. The NAVI subjects are
small and largely convex, so one view already fixes their appearance and the
remaining references buy cross-view agreement rather than identity. ViCLIP is
saturated for every count on both benchmarks, as prompt adherence does not
depend on how many views describe the subject.

\section{Training Data Comparison}
\label{app:traindata}

Figure~\ref{fig:datacompare} contrasts our renders with the output of a
camera-controllable image-to-video generator qualitatively. Here we measure the
same contrast. We pair every clip of our corpus with the Uni3C~\citep{cao2025uni3c}
clip generated for the same subject and camera path, and we score both with the
metrics of Section~\ref{subsec:setup}, using the same four reference views for
reference-based metrics and the same source mesh for GT-CD. A clip
that is geometrically faithful to the subject should score as a well-behaved
orbit under these metrics regardless of how it was produced.
Table~\ref{tab:st_traindata} reports the outcome.

\begin{table}[h]
\centering
\small
\setlength{\tabcolsep}{8pt}
\renewcommand{\arraystretch}{1.15}
\caption{\textbf{Training clips scored as generations.} Paired clips of the same
subject under the same camera path and ground-truth geometry. Best results are
in \textbf{bold}.}
\label{tab:st_traindata}
\begin{tabular}{l ccccc}
\toprule
Source & DINO$\uparrow$ & CLIP$\uparrow$ & MEt3R$\downarrow$ & GT-CD$\downarrow$ & RotCov$\downarrow$ \\
\midrule
Uni3C~\citep{cao2025uni3c}    & 0.7167 & 0.9191 & 0.3052 & 0.1409 & 0.6661 \\
\textbf{Ours (3D-asset render)} & \textbf{0.8030} & \textbf{0.9385} & \textbf{0.1871} & \textbf{0.1323} & \textbf{0.6318} \\
\bottomrule
\end{tabular}
\end{table}

\section{Baselines Extension}
\label{app:baseline-ext}

Phantom and MAGREF are single-view subject-to-video models. For the comparison
in Table~\ref{tab:main} we extend them to accept all four reference views
following MV-S2V~\citep{song2026mvs2v}, which is the setting their multi-view
rows report. Here we also run them in their
original single-view mode, where only the first reference view enters the model
and the prompt is unchanged, so that the effect of the extension is visible on
its own. Everything else, the checkpoint, the sampler, the trajectory sentence
and the evaluation protocol, is held fixed. Tables~\ref{tab:st_phantom_sv}
and~\ref{tab:st_magref_sv} report the two settings side by side.

\begin{table}[h]
\centering
\small
\setlength{\tabcolsep}{5pt}
\renewcommand{\arraystretch}{1.15}
\caption{\textbf{Phantom with one reference view and with four.} We compare
single-view input with the four-view setting used in Table~\ref{tab:main}.
GT-CD uses ground-truth geometry. Best results
per benchmark are in \textbf{bold}.}
\label{tab:st_phantom_sv}
\resizebox{\linewidth}{!}{%
\begin{tabular}{l ccccc ccccc}
\toprule
 & \multicolumn{5}{c}{Held-out} & \multicolumn{5}{c}{NAVI} \\
\cmidrule(lr){2-6}\cmidrule(lr){7-11}
Input
 & DINO$\uparrow$ & CLIP$\uparrow$ & MEt3R$\downarrow$ & GT-CD$\downarrow$ & RotCov$\downarrow$
 & DINO$\uparrow$ & CLIP$\uparrow$ & MEt3R$\downarrow$ & GT-CD$\downarrow$ & RotCov$\downarrow$ \\
\midrule
Single view (original) & 0.5921 & 0.8609 & 0.4736 & \textbf{0.1651} & 0.9254 & 0.6483 & 0.8677 & 0.5058 & \textbf{0.1791} & 1.4705 \\
Four views (reported)  & \textbf{0.621} & \textbf{0.871} & \textbf{0.444} & 0.2843 & \textbf{0.906} & \textbf{0.690} & \textbf{0.882} & \textbf{0.462} & 0.2014 & \textbf{1.458} \\
\bottomrule
\end{tabular}}
\end{table}

\begin{table}[h]
\centering
\small
\setlength{\tabcolsep}{5pt}
\renewcommand{\arraystretch}{1.15}
\caption{\textbf{MAGREF with one reference view and with four.} We compare
single-view input with the four-view setting used in Table~\ref{tab:main}.
GT-CD uses ground-truth geometry. Best results per benchmark are in
\textbf{bold}.}
\label{tab:st_magref_sv}
\resizebox{\linewidth}{!}{%
\begin{tabular}{l ccccc ccccc}
\toprule
 & \multicolumn{5}{c}{Held-out} & \multicolumn{5}{c}{NAVI} \\
\cmidrule(lr){2-6}\cmidrule(lr){7-11}
Input
 & DINO$\uparrow$ & CLIP$\uparrow$ & MEt3R$\downarrow$ & GT-CD$\downarrow$ & RotCov$\downarrow$
 & DINO$\uparrow$ & CLIP$\uparrow$ & MEt3R$\downarrow$ & GT-CD$\downarrow$ & RotCov$\downarrow$ \\
\midrule
Single view (original) & 0.5804 & 0.865 & 0.4538 & \textbf{0.1752} & 1.1974 & 0.657 & 0.8729 & \textbf{0.4529} & \textbf{0.1704} & \textbf{1.2712} \\
Four views (reported)  & \textbf{0.616} & \textbf{0.876} & \textbf{0.409} & 0.1789 & \textbf{1.177} & \textbf{0.665} & \textbf{0.892} & 0.457 & 0.2145 & 1.284 \\
\bottomrule
\end{tabular}}
\end{table}

\section{Dataset}
\label{app:dataset}

\subsection{Curation and composition}
\label{app:funnel}

\textsc{Showcase-4K} is the surviving remainder of a much larger candidate pool.
Assets enter from Objaverse~\citep{deitke2023objaverse}, the Digital Twin
Catalog~\citep{dong2025dtc} and Google Scanned Objects~\citep{downs2022gso} and
pass through four gates: \textbf{G0}, a model-free heuristic on the mesh and a
quick render; \textbf{G1}, a vision-language check of four rendered views;
\textbf{G2}, the same check on the composited target video; and \textbf{G3},
on the reconstructed turntable (rubrics in Appendix~\ref{app:rubric_gates}).
G0 rejected $17\%$ of downloaded candidates outright, and G1 kept $4{,}186$ of
the $7{,}016$ assets it judged ($60\%$). In the most recent production round,
$44\%$ of $7{,}438$ composited clips passed G2 and $63\%$ of those passed G3,
so $28\%$ of rendered clips survive; most G2 rejections are a changed object
structure or physically implausible support. The binding constraint is G2, and
we did not relax it. Over-generating and discarding is cheaper than letting a
bad sample into training, because a subject that deforms in the target teaches
exactly the failure the method is meant to remove. Over all rounds $4{,}829$
clips passed, and $4{,}000$ of them, covering $2{,}593$ assets in $620$
categories, form the released corpus.

Curation runs without a human in the loop: every semantic judgement is made by
Gemini~2.5~Flash, which returns a binary score and a short rationale per
criterion. Over all rounds the gates took about $29{,}000$ calls at about
$120$~USD, roughly $25$~USD per thousand kept clips, so expanding the corpus
scales with that cost rather than with annotator time.

\subsection{Camera trajectories}
\label{sec:traj-tags}

Every prompt ends with one sentence describing the camera path. To keep the
figure captions short we label each clip with a tag instead, and give here the
sentence each tag stands for. A tag is written as a motion followed by either a
rotation amount (Orbit and Helix) or a framing transition (everything else), for example
Helix-Up 360\textdegree{} or Arc-Down L$\rightarrow$R. The corpus covers every motion
below, with the Orbit and Helix motions contributing the object-centric half and the arc, truck,
dolly and zoom motions the extended-trajectory half.

The two suffix families do not mix. The four orbit-family motions (Orbit and Helix-Up/Down/In) each take one of four
rotation amounts, which gives $16$ orbit-family trajectories, and the remaining eight
motions each take one of nine framing transitions, which gives $72$ more, so the
vocabulary defines $88$ distinct camera trajectories in total.

\begin{figure}[h]
    \centering
    \includegraphics[width=0.92\linewidth]{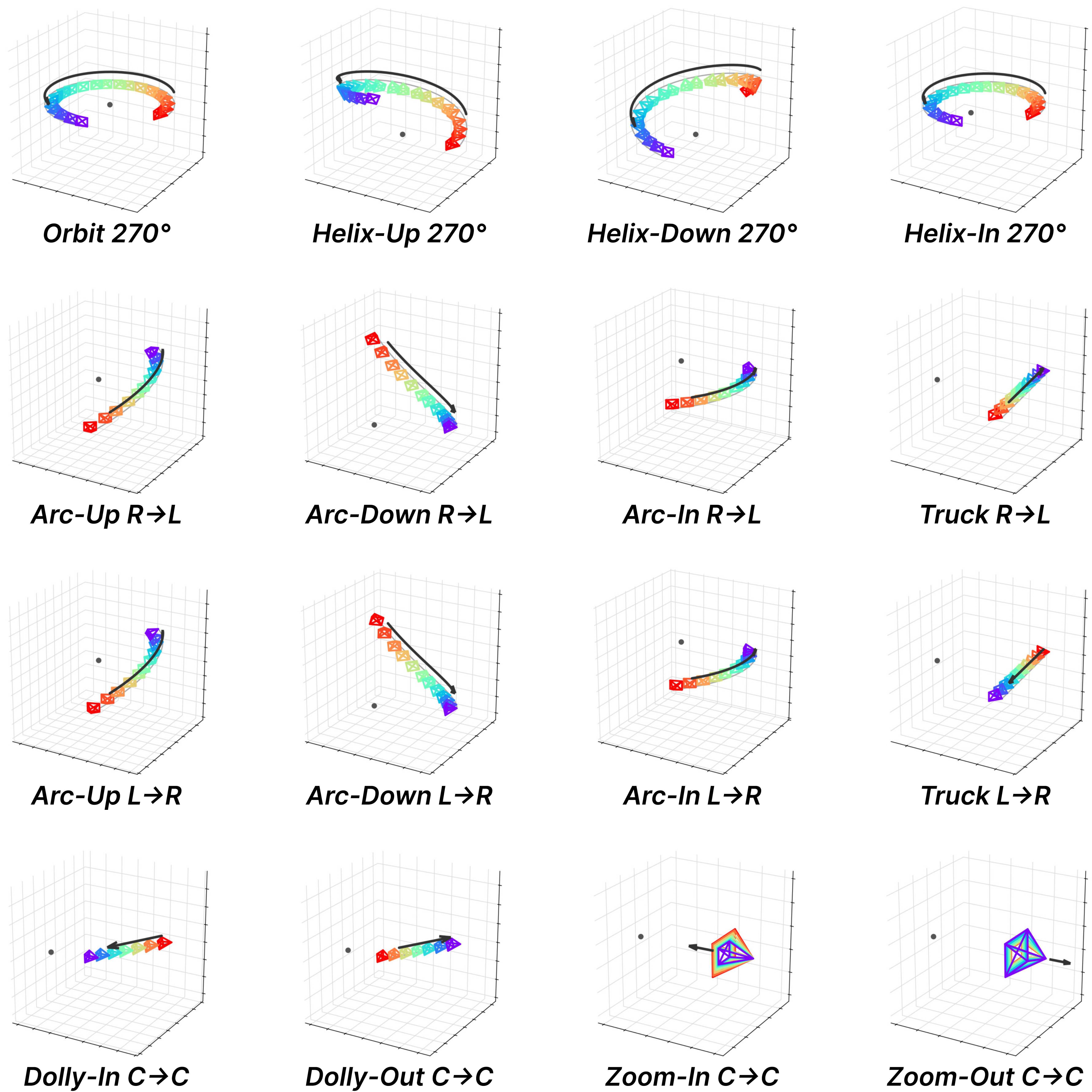}
    \caption{\textbf{Camera trajectories.} Sixteen representative trajectories
    out of the $88$ the vocabulary defines, one per tag family. Each panel draws
    the camera frusta sampled along the path, with the arrow giving the direction
    of travel and the grey dot marking the subject.}
    \label{fig:traj}
\end{figure}

Table~\ref{tab:traj-tags} lists the motions, and Tables~\ref{tab:traj-rotation}
and~\ref{tab:traj-framing} list the two suffixes.
Figure~\ref{fig:traj} draws sixteen of the $88$ as camera paths.

\begin{table}[h]
\centering
\small
\setlength{\tabcolsep}{4pt}
\renewcommand{\arraystretch}{1.15}
\caption{\textbf{Motion vocabulary.} The sentence each motion contributes to
the prompt. For Orbit and Helix the amount is filled from Table~\ref{tab:traj-rotation}.
Every other motion is followed by a framing fragment from
Table~\ref{tab:traj-framing}.}
\label{tab:traj-tags}
\begin{tabular}{@{}l p{0.80\linewidth}@{}}
\toprule
Tag & Sentence \\
\midrule
Orbit      & The camera circles \emph{[amount]} around the object. \\
Helix-Up   & The camera circles \emph{[amount]} around the object while rising. \\
Helix-Down & The camera circles \emph{[amount]} around the object while descending. \\
Helix-In   & The camera circles \emph{[amount]} around the object while moving closer. \\
Arc-Up     & The camera circles partway around the object, rising from eye level to look down on it while moving closer, ending in front \\
Arc-Down   & The camera starts high above the object and comes down to eye level while circling partway around and moving closer, ending in front \\
Arc-In     & The camera stays at eye level and circles partway around the object while moving closer, ending in a tight front view \\
Truck      & The camera glides sideways past the object at eye level, holding its distance as the object sweeps through the frame \\
Dolly-In   & The camera moves straight toward the object without circling, filling the frame \\
Dolly-Out  & The camera moves straight back from the object without circling, revealing the space around it \\
Zoom-In    & The camera stays in place and zooms in, narrowing the field of view \\
Zoom-Out   & The camera stays in place and zooms out, widening the field of view \\
\bottomrule
\end{tabular}
\end{table}

\begin{table}[h]
\centering
\small
\setlength{\tabcolsep}{4pt}
\renewcommand{\arraystretch}{1.15}
\caption{\textbf{Rotation amount.} Arc motions carry their rotation in the
trajectory definition and take no suffix. Truck, Dolly and Zoom do not rotate.}
\label{tab:traj-rotation}
\begin{tabular}{@{}ll@{}}
\toprule
Suffix & Sentence fragment \\
\midrule
90\textdegree{}  & \ldots{}circles a quarter of the way around the object \\
180\textdegree{} & \ldots{}circles halfway around the object \\
270\textdegree{} & \ldots{}circles three-quarters of the way around the object \\
360\textdegree{} & \ldots{}circles all the way around the object \\
\bottomrule
\end{tabular}
\end{table}

\begin{table}[h]
\centering
\small
\setlength{\tabcolsep}{4pt}
\renewcommand{\arraystretch}{1.15}
\caption{\textbf{Framing transition.} Where the object starts and ends inside
the frame, with L, C and R for the left, center and right. The same fragment is
appended regardless of the motion.}
\label{tab:traj-framing}
\begin{tabular}{@{}ll@{}}
\toprule
Suffix & Sentence fragment \\
\midrule
L$\rightarrow$L & with the object held on the left of the frame throughout \\
L$\rightarrow$C & with the object starting on the left of the frame and settling into the center \\
L$\rightarrow$R & with the object starting on the left of the frame and drifting across to the right \\
C$\rightarrow$L & with the object starting at the center of the frame and drifting across to the left \\
C$\rightarrow$C & with the object held at the center of the frame throughout \\
C$\rightarrow$R & with the object starting at the center of the frame and drifting across to the right \\
R$\rightarrow$L & with the object starting on the right of the frame and drifting across to the left \\
R$\rightarrow$C & with the object starting on the right of the frame and settling into the center \\
R$\rightarrow$R & with the object held on the right of the frame throughout \\
\bottomrule
\end{tabular}
\end{table}

\section{Failure Analysis}
\label{app:failure}

\begin{figure}[h]
    \centering
    \includegraphics[width=\linewidth]{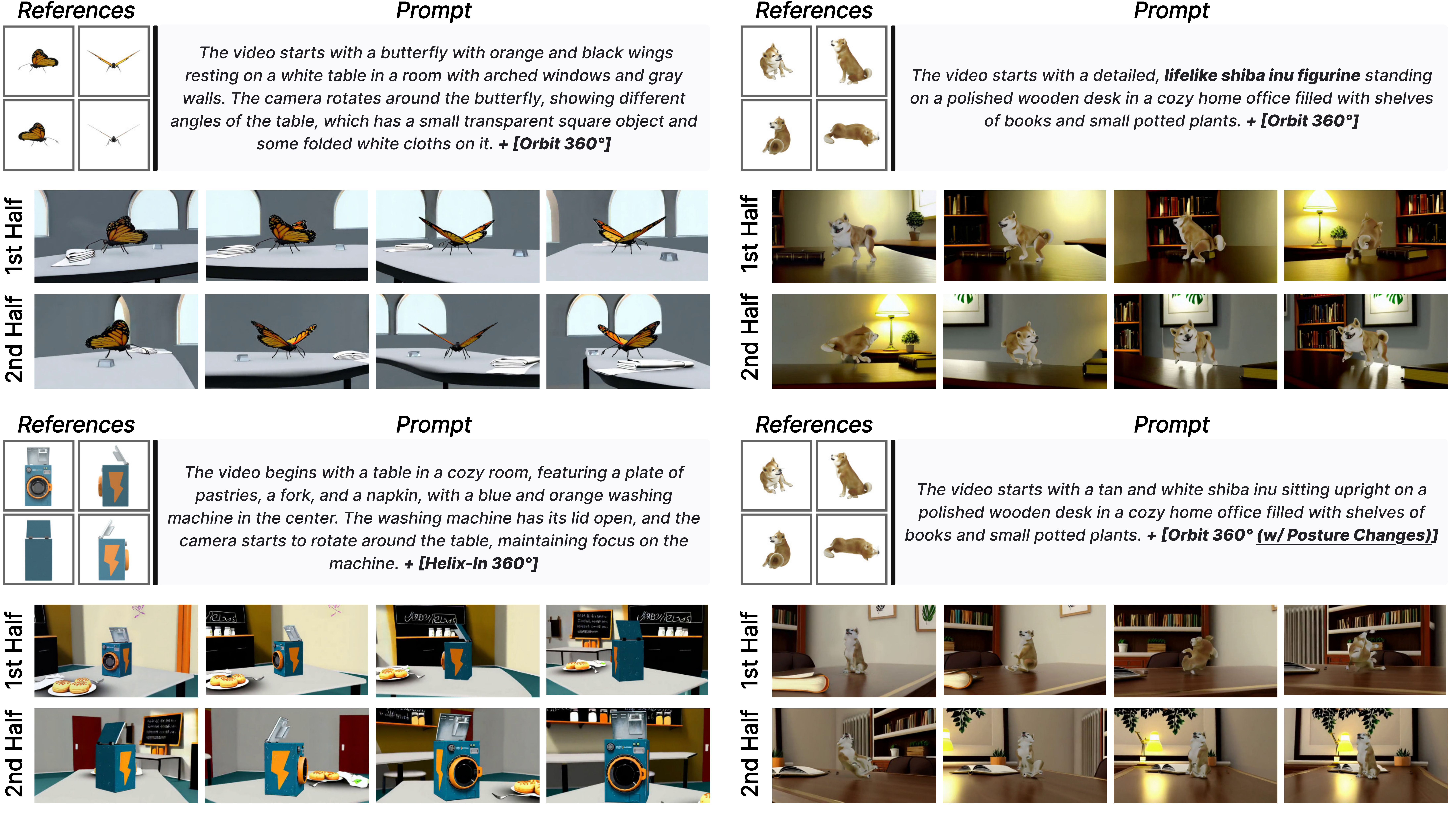}
    \caption{\textbf{Failure cases.} \emph{Left two.} Prompt adherence and the
    commanded rotation both hold, but the surroundings render with flat shading
    and little material detail, so the clip reads as slightly synthetic.
    \emph{Right two.} Both receive four reference views of a subject caught in
    four different poses. The upper one is prompted in the usual static style
    and settles into an averaged pose that matches none of the references, while
    the lower one is asked to follow the motion in step with the orbit and
    tangles in 3D at the transitions between poses.}
    \label{fig:sf_failure}
\end{figure}

\paragraph{Analysis.}
\label{app:failure-analysis}
The two columns of Figure~\ref{fig:sf_failure} fail for unrelated reasons. The
synthetic look of the background follows from how the training targets are
built. Only the subject is physically rendered, and the scene around it is
painted in afterwards by a video inpainting model conditioned on the prompt. The corpus
therefore pairs a physically shaded subject with a generated background, and the
light transport between the two was never consistent to begin with. The model
reproduces that statistic, so the subject holds while the surroundings inherit
the flat shading and thin material detail of their source.

The dynamic case is a property of the conditioning signal itself. The
surface-normal turntable is obtained by lifting the reference views into a
single static reconstruction and rendering that one shape from every
viewpoint. When the four views catch the subject in four different poses, the
reconstruction fuses them into one rigid body that corresponds to no pose the
subject ever held, and the condition handed to the generator is a static
representation of a subject that is not static. Under a static prompt the model
follows that condition faithfully, which is exactly why the output settles into
an averaged pose. Asking for motion in the prompt does not repair it. The text
now requests a pose change that the geometry condition does not encode, so the
two disagree, and the model has to interpolate between reference poses in image
space with no 3D correspondence to guide it. The tangles appear at those
interpolated transitions. Handling non-rigid subjects would require a
conditioning representation that varies over the generated sequence, which we
leave to future work.

\section{Prompts and Rubrics}
\label{app:rubric}

This section describes the final pipeline configuration: the curation
criteria, category and scene authoring rules, captioning templates, and
inference prompt format (Table~\ref{tab:st_prompts}). The gate criteria are
presented with shared output formatting. The quantitative evaluation in
Section~\ref{subsec:setup} uses no LLM judge.

\begin{table}[h]
\centering\footnotesize
\caption{\textbf{Where a language or vision-language model is used.}
Each gate accepts a sample only if all of its criteria score $1$.
Gate calls use temperature $0$.}
\label{tab:st_prompts}
\setlength{\tabcolsep}{5pt}
\renewcommand{\arraystretch}{1.15}
\begin{tabularx}{\linewidth}{@{}l l >{\raggedright\arraybackslash}X >{\raggedright\arraybackslash}X@{}}
\toprule
Stage & Model & Input & Output \\
\midrule
G1 asset gate & Gemini 2.5 Flash & 4 rendered views, category label & 6 binary criteria \\
G2 target gate & Gemini 2.5 Flash & 4 reference views, 16 target frames & 6 binary criteria \\
G3 geometry gate & Gemini 2.5 Flash & 4 reference views, 16 RGB reconstruction frames & 4 binary criteria \\
Category descriptions & Claude Opus 5 & category label & appearance sentence, coarse type \\
Scene vocabulary & Claude Opus 5 & --- & scene bank (2,000 locations) \\
Captioning & Tarsier2-7B & video & condition and target captions \\
\bottomrule
\end{tabularx}
\end{table}

\subsection{Curation gates}
\label{app:rubric_gates}

\textbf{G0} is a model-free heuristic that screens meshes and quick renders
for degeneracy. \textbf{G1}--\textbf{G3} use the criteria below. G2 and G3
receive $16$ uniformly sampled video frames; G3 evaluates the RGB rendering
of the reconstruction. For non-orbit trajectories, the same six G2 criteria
are worded for general camera motion and off-center framing.

All three gates return a binary score and a short rationale for each
criterion, together with an overall acceptance decision, in JSON format.
A sample passes a gate only when all of its criteria score $1$.

\begin{promptbox}{G1: Asset Quality}
\par\medskip\noindent{\footnotesize\bfseries\scshape Prompt}\par\nopagebreak\noindent
\par\noindent You are an asset QC reviewer for a 3D object dataset. I show reference render views (view0-3) of ONE 3D asset plus its category label. The bar is "good enough to show in a paper/demo" - NOT perfection. Score 6 criteria 0/1.
\par\medskip\noindent{\footnotesize\bfseries\scshape Criteria} (score each 0/1 with a short justification)\par\nopagebreak
\begin{list}{}{\leftmargin=2.1em \labelwidth=1.7em \labelsep=0.4em \itemsep=4pt \parsep=0pt \topsep=2pt \partopsep=0pt}
\item[(1)] \textbf{realism}
\par\noindent (read this as ASSET CRAFTSMANSHIP, not "is the subject a real thing"): judge how WELL-MADE the 3D asset is.
\par\noindent\hangindent=1.2em\hangafter=1 - Being a figurine, plush toy, statue, or a stylized animal model is NOT a defect - many categories are inherently toy-like, and a well-crafted figurine is a good asset.
\par\noindent\hangindent=1.2em\hangafter=1 - The decisive test is MATERIAL AND DETAIL: a good asset has believable surface material (fabric weave, fur, metal, glaze, wood grain, paint with wear) and real modelled detail.
\par\noindent\hangindent=1.2em\hangafter=1 - Score 0 (reject) when the asset is low-effort or cartoon-flat: emoji-like or clip-art forms with flat uniform fills, crude blocky/faceted shapes, plastic-smooth surfaces with no material character, sloppy or melted geometry, or a shape so distorted you cannot tell what it is.
\par\noindent\hangindent=1.2em\hangafter=1 - Score 1 (keep) when the craftsmanship is solid - including figurines, sculpted animals, plush toys and other inherently toy-like SUBJECTS - as long as the asset itself shows real material and detail.
\par\noindent\hangindent=1.2em\hangafter=1 - So: a finely sculpted animal figurine = 1; a flat cartoon hamburger or a toy-car with uniform purple plastic = 0.
\item[(2)] \textbf{geometry}
\par\noindent proper, complete geometry - NO low-poly/blocky faceting, holes, broken or degenerate mesh.
\item[(3)] \textbf{no\_base\_plate}
\par\noindent LOOK AT WHAT THE OBJECT RESTS ON.
\par\noindent\hangindent=1.2em\hangafter=1 - Score 0 if the asset includes any surface/terrain the object merely sits on and that would not be carried with it: a scanned ground or dirt patch, sand/water/grass terrain, baked-in cast-shadow geometry, a cake board, a serving plate, a square showroom slab, or a fragment of surrounding wall or floor.
\par\noindent\hangindent=1.2em\hangafter=1 - A base that is an INTEGRAL part of the real object - a trophy's plinth, a flower pot, a lamp base, a statue's pedestal, a vase foot - is NOT a violation (score 1).
\item[(4)] \textbf{texture}
\par\noindent has a proper material/color appearance - NOT a blank untextured grey/white placeholder mesh or an unfinished flat single-color surface.
\par\noindent\hangindent=1.2em\hangafter=1 - Smooth surfaces are FINE (car paint, metal kettle, glass) - score 0 only when it looks like an untextured/unfinished mesh.
\item[(5)] \textbf{category\_match}
\par\noindent score 0 if (a) the object is plainly a DIFFERENT KIND of thing than the label, or (b) the asset is not ONE instance of it - multiple copies bundled together (three bats), or the object plus surrounding structure / a mini-scene (a phone booth attached to a wall slab).
\par\noindent\hangindent=1.2em\hangafter=1 - Judge kind at the COARSE level: close relatives, subspecies, breeds, and stylized variants COUNT AS A MATCH (a wolf-like dog for 'wolf', any motorcycle for 'dirt\_bike', any van for 'minivan').
\par\noindent\hangindent=1.2em\hangafter=1 - Do not fail on fine-grained species/model distinctions.
\item[(6)] \textbf{standing\_thin}
\par\noindent score 0 ONLY in the specific failure case where the object is a THIN, PLATE-LIKE OR LOOP-LIKE form standing ON ITS EDGE, so that rotating 90 degrees makes its silhouette collapse to a narrow sliver - a ring or coin balanced on edge, an upright bat or pole, a hanging necklace/chain.
\par\noindent\hangindent=1.2em\hangafter=1 - Compare the 4 views: if one view is a full shape and the neighbouring view is a thin line, score 0.
\par\noindent\hangindent=1.2em\hangafter=1 - If the object has real volume from every angle (an animal, a vehicle, a cup, a shoe, a fish lying horizontally), score 1 even if it is somewhat slender.
\par\noindent\hangindent=1.2em\hangafter=1 - When unsure, score 1.
\end{list}
\par\medskip\noindent{\footnotesize\bfseries\scshape Rule}\par\nopagebreak\noindent{}qualified only if ALL 6 = 1.
\end{promptbox}

\begin{promptbox}{G2: Target Video Quality}
\par\medskip\noindent{\footnotesize\bfseries\scshape Prompt}\par\nopagebreak\noindent
\par\noindent You are a QC reviewer for an object-centric TURNTABLE video dataset. The camera ORBITS a central object in a fixed scene.
\par\noindent I will give you, IN ORDER: (1) the REAL ORIGINAL object as reference photos (view0-3) - ground truth; (2) the GENERATED video frames to CHECK.
\par\noindent The CAMERA orbits while the object stays FIXED in the scene, so seeing different ANGLES of the SAME place is EXPECTED - do NOT penalize that. But if the background is nearly STATIC while the object SPINS in place, that is a turntable spin, not a camera orbit - fail physical\_plausibility.
\par\medskip\noindent{\footnotesize\bfseries\scshape Criteria} (score each 0/1 with a short justification)\par\nopagebreak
\begin{list}{}{\leftmargin=2.1em \labelwidth=1.7em \labelsep=0.4em \itemsep=4pt \parsep=0pt \topsep=2pt \partopsep=0pt}
\item[(1)] \textbf{authenticity}
\par\noindent object grounded, NOT floating.
\item[(2)] \textbf{background\_sharpness}
\par\noindent the whole scene stays in focus.
\par\noindent\hangindent=1.2em\hangafter=1 - Score 0 if the background is blurred or bokeh'd into mush while only the object is sharp.
\item[(3)] \textbf{scene\_consistency}
\par\noindent the background must behave like ONE RIGID 3D place.
\par\noindent\hangindent=1.2em\hangafter=1 - Score 0 if structures appear/vanish between frames, the layout rearranges, floor/tile grids warp, or it becomes a DIFFERENT place.
\par\noindent\hangindent=1.2em\hangafter=1 - Seeing NEW parts of the same room as the camera moves is fine.
\item[(4)] \textbf{physical\_plausibility}
\par\noindent no non-physical deformation/flicker/fusion of the central object, and the object must be physically SUPPORTED by what it rests on - not floating, not in an impossible pose, not on a surface that could not hold it.
\item[(5)] \textbf{subject\_shape}
\par\noindent object is not paper/card/cloth/2D-planar.
\item[(6)] \textbf{object\_integrity}
\par\noindent FIRST, from the REFERENCE photos, note the original object's parts (strap? pole/stand? flag? handle?).
\par\noindent\hangindent=1.2em\hangafter=1 - THEN check the GENERATED frames.
\par\noindent\hangindent=1.2em\hangafter=1 - Score 0 ONLY for a CLEAR STRUCTURAL change: a substantial new part attached to the object (a pedestal or stand under it, a frame around it), the object FUSED INTO a background structure (a bird becoming a fountain pillar), or its overall shape MORPHED into a different form.
\par\noindent\hangindent=1.2em\hangafter=1 - Do NOT fail on: parts that merely become visible from a new camera angle, small texture noise or thin stray artifacts, lighting and shadow differences, or nearby scene props that sit SEPARATELY next to the object without touching it.
\par\noindent\hangindent=1.2em\hangafter=1 - When the object still reads as the same item with the same silhouette, score 1.
\end{list}
\par\medskip\noindent{\footnotesize\bfseries\scshape Rule}\par\nopagebreak\noindent{}qualified only if ALL 6 = 1.
\pfield{Final user instruction}Now evaluate. Any part in GENERATED but absent in REFERENCE = object\_integrity 0.
\end{promptbox}

G3 is used only for training-data curation and is not applied during evaluation or inference.

\begin{promptbox}{G3: Reconstruction Quality}
\par\medskip\noindent{\footnotesize\bfseries\scshape Prompt}\par\nopagebreak\noindent
\par\noindent You are a QC reviewer for 3D RECONSTRUCTION quality. A reconstruction of an object is rendered as a turntable video (camera orbits; the object should appear to spin in place). I will give you, IN ORDER: (1) the REAL ORIGINAL object as reference photos (view0-3) - ground truth; (2) the RECONSTRUCTION frames to CHECK.
\par\noindent Judge ONLY the reconstructed object's GEOMETRY vs the reference. Object SIZE and POSITION in the frame are handled elsewhere - do NOT penalize framing/crop/margins.
\par\medskip\noindent{\footnotesize\bfseries\scshape Criteria} (score each 0/1 with a short justification)\par\nopagebreak
\begin{list}{}{\leftmargin=2.1em \labelwidth=1.7em \labelsep=0.4em \itemsep=4pt \parsep=0pt \topsep=2pt \partopsep=0pt}
\item[(1)] \textbf{identity\_match}
\par\noindent compare the object's OVERALL PROPORTIONS / aspect ratio AND distinctive 3D FORM to the reference.
\par\noindent\hangindent=1.2em\hangafter=1 - 0 if proportions are distorted (e.g. a wide rectangular object reconstructed as a square/cube), if melted/morphed, or if reduced to a generic blob that lost the reference's distinctive form.
\par\noindent\hangindent=1.2em\hangafter=1 - IGNORE surface text/printed labels/logos - MV-SAM3D cannot reconstruct fine print, so blurred/changed/flickering text is NOT a defect; judge by 3D shape and proportion only.
\item[(2)] \textbf{completeness}
\par\noindent object is whole - no holes, no missing or collapsed STRUCTURAL parts.
\par\noindent\hangindent=1.2em\hangafter=1 - A flickering/blurred printed LABEL or surface text is NOT a completeness defect (surface print is expected to degrade); only missing 3D parts count.
\item[(3)] \textbf{no\_floater}
\par\noindent no floating fragments, stray blobs, or disconnected debris around the object.
\item[(4)] \textbf{rotation\_integrity}
\par\noindent object spins IN PLACE about its own axis.
\par\noindent\hangindent=1.2em\hangafter=1 - 0 if it SWINGS/ORBITS around an off-center point (revolves).
\end{list}
\par\medskip\noindent{\footnotesize\bfseries\scshape Rule}\par\nopagebreak\noindent{}qualified only if ALL 4 = 1.
\end{promptbox}

\subsection{Category descriptions and scene vocabulary}
\label{app:rubric_other}

Category descriptions and scene vocabulary are generated automatically
with Claude Opus 5 under the rules below. Each category has an appearance
sentence and a coarse type; the type selects compatible scene locations.
The scene bank contains $2{,}000$ locations with surfaces, props, depth cues
and lighting used to assemble the compositing prompts
(Section~\ref{subsec:method_data}).

\begin{promptbox}{Category Descriptions and Scene Vocabulary}
\pfield{Category description}one sentence per category, written from the category name (not from a particular asset):
\par\noindent\hspace*{1.5em}"A [adjective] [material] <category> with [distinguishing trait]"
\par\noindent\hangindent=1.2em\hangafter=1 - plus exactly one coarse type from \{accessory, apparel, appliance, bag, beauty, decor, drink, electronics, food, furniture, sports, tools\}
\pfield{Scene vocabulary}one entry per location: a compatible set of surfaces, props, depth cues and lighting, with a tone label (cool, warm, neutral, green).
\par\noindent\hangindent=1.2em\hangafter=1 - no words implying defocus or bokeh
\par\noindent\hangindent=1.2em\hangafter=1 - no people or actions
\par\noindent\hangindent=1.2em\hangafter=1 - no description of the subject itself
\end{promptbox}

\subsection{Captioning}
\label{app:rubric_caption}

Tarsier2-7B captions the RGB reconstruction turntable and the composited
target with greedy decoding and a category hint. Before insertion into
\texttt{[CONDITION]}, the reconstruction caption is stripped of turntable
narration and references to reconstruction and the white background.
The condition box below shows the resulting format; the target box gives
the captioning prompt.

\begin{promptbox}{Condition Caption Format}
\pfield{Stripping rule}Retain the object's visible appearance and distinctive parts. Remove turntable-rotation narration, white-background phrases and reconstruction-specific narration.
\pfield{Template}"The video starts with [object: shape, color, material, distinctive visible features]. [Optional additional appearance details.]"
\pfield{Example}"The video starts with a black gargoyle statue with bat-like wings and a menacing expression, positioned on a stone base."
\pfield{Output}The stripped object description is used after [CONDITION].
\end{promptbox}

\begin{promptbox}{Target Caption Prompt}
\pfield{Prompt}In this video, a 3D object is placed in a real-world scene. Focus your description on the object, where it is placed, the surrounding scene, and any visible action or scene change.
\par\smallskip\noindent Describe the video in 2 to 3 sentences. The first sentence MUST start with "The video starts with" and contain the object, its placement, and the scene. Follow this structure precisely:
\pfield{Template}"The video starts with [object: shape, color, material, distinctive features] [placement phrase: e.g., placed on a wooden table / standing on a stone pedestal / resting on a tiled floor] in [scene location with lighting style]. [Optional second/third sentence: brief description of camera motion, scene change, or other notable action visible in the video]."
\pfield{Example}"The video starts with a digital watch with a metal bracelet placed on a rusty riveted metal table in a brick-walled loft with warm natural light from arched windows. The camera gradually moves closer to the watch, revealing the textured brick walls and wooden floor in more detail."
\pfield{Output}Respond ONLY with the description.
\end{promptbox}

\subsection{Inference prompts}
\label{app:rubric_prompts}

The captioner does not reliably state how far the camera turns, so a fixed
sentence derived from the camera mode is appended to every target caption. At
inference our model receives an introductory sentence, the stripped condition caption
after \texttt{[CONDITION]}, and the target caption after \texttt{[VIDEO]},
ending with the trajectory sentence; baselines receive only the text after
\texttt{[VIDEO]}.

\begin{promptbox}{Inference Prompt}
\pfield{Trajectory sentence}Chosen from the motion and suffix vocabulary in Tables~\ref{tab:traj-tags}--\ref{tab:traj-framing} and appended to the target caption.
\pfield{Structure}
\par\noindent\hspace*{1.5em}intro sentence
\par\noindent\hspace*{1.5em}+ [CONDITION] stripped condition caption
\par\noindent\hspace*{1.5em}+ [VIDEO] target caption
\par\noindent\hspace*{1.5em}+ trajectory sentence
\pfield{Example}This video demonstrates a multi-view capture of a gargoyle, showcasing the object as it rotates and moves across different viewpoints.
\par\noindent [CONDITION] The video starts with a black gargoyle statue with bat-like wings and a menacing expression, positioned on a stone base.
\par\noindent [VIDEO] The video begins with a detailed view of a gargoyle statue situated on a stone base, surrounded by scattered autumn leaves on a cracked stone pavement, with a gothic architectural background featuring tall spires and intricate stonework. The camera rotates around the gargoyle, capturing different angles and perspectives [...] The camera circles all the way around the object while rising.
\pfield{Baselines}the text after [VIDEO] only; for 3DreamBooth it is rewritten into its identifier format: "A video of a [rhs <noun>] ..."
\end{promptbox}

%% file: main.bib
@String(CVPR= {IEEE Conf. Comput. Vis. Pattern Recog.})

@String(ICCV= {Int. Conf. Comput. Vis.})

@String(ECCV= {Eur. Conf. Comput. Vis.})

@String(NIPS= {Adv. Neural Inform. Process. Syst.})

@String(TOG= {ACM Trans. Graph.})

@String(ICLR = {Int. Conf. Learn. Represent.})

@String(CVPR  = {CVPR})

@String(ICCV  = {ICCV})

@String(ECCV  = {ECCV})

@String(NIPS  = {NeurIPS})

@String(TOG   = {ACM TOG})

@String(ICLR  = {ICLR})

@inproceedings{ho2020ddpm,
  author = {Jonathan Ho and Ajay Jain and Pieter Abbeel},
  title = {Denoising Diffusion Probabilistic Models},
  booktitle = NIPS, year = 2020}

@inproceedings{rombach2022ldm,
  author = {Robin Rombach and Andreas Blattmann and Dominik Lorenz and Patrick Esser and Bj{\"o}rn Ommer},
  title = {High-Resolution Image Synthesis with Latent Diffusion Models},
  booktitle = CVPR, year = 2022}

@inproceedings{peebles2023dit,
  author = {William Peebles and Saining Xie},
  title = {Scalable Diffusion Models with Transformers},
  booktitle = ICCV, year = 2023}

@inproceedings{esser2024sd3,
  author = {Patrick Esser and Sumith Kulal and Andreas Blattmann and Rahim Entezari and Jonas M{\"u}ller and Harry Saini and others},
  title = {Scaling Rectified Flow Transformers for High-Resolution Image Synthesis},
  booktitle = {ICML}, year = 2024}

@article{blattmann2023svd,
  author = {Andreas Blattmann and Tim Dockhorn and Sumith Kulal and Daniel Mendelevitch and Maciej Kilian and Dominik Lorenz and others},
  title = {Stable Video Diffusion: Scaling Latent Video Diffusion Models to Large Datasets},
  journal = {arXiv preprint arXiv:2311.15127}, year = 2023}

@inproceedings{guo2024animatediff,
  author = {Yuwei Guo and Ceyuan Yang and Anyi Rao and Zhengyang Liang and Yaohui Wang and Yu Qiao and Maneesh Agrawala and Dahua Lin and Bo Dai},
  title = {AnimateDiff: Animate Your Personalized Text-to-Image Diffusion Models without Specific Tuning},
  booktitle = ICLR, year = 2024}

@inproceedings{yang2025cogvideox,
  author = {Zhuoyi Yang and Jiayan Teng and Wendi Zheng and Ming Ding and Shiyu Huang and Jiazheng Xu and others},
  title = {CogVideoX: Text-to-Video Diffusion Models with an Expert Transformer},
  booktitle = ICLR, year = 2025}

@article{kong2024hunyuanvideo,
  author = {Weijie Kong and Qi Tian and Zijian Zhang and Rox Min and Zuozhuo Dai and Jin Zhou and others},
  title = {HunyuanVideo: A Systematic Framework for Large Video Generative Models},
  journal = {arXiv preprint arXiv:2412.03603}, year = 2024}

@article{wan2025,
  author = {{Wan Team}},
  title = {Wan: Open and Advanced Large-Scale Video Generative Models},
  journal = {arXiv preprint arXiv:2503.20314}, year = 2025}

@misc{brooks2024sora,
  author = {Tim Brooks and Bill Peebles and Connor Holmes and Will DePue and Yufei Guo and Li Jing and others},
  title = {Video Generation Models as World Simulators},
  howpublished = {\url{https://openai.com/research/video-generation-models-as-world-simulators}}, year = 2024}

@inproceedings{singer2023makeavideo,
  author = {Uriel Singer and Adam Polyak and Thomas Hayes and Xi Yin and Jie An and Songyang Zhang and others},
  title = {Make-A-Video: Text-to-Video Generation without Text-Video Data},
  booktitle = ICLR, year = 2023}

@inproceedings{mildenhall2020nerf,
  author = {Ben Mildenhall and Pratul P. Srinivasan and Matthew Tancik and Jonathan T. Barron and Ravi Ramamoorthi and Ren Ng},
  title = {NeRF: Representing Scenes as Neural Radiance Fields for View Synthesis},
  booktitle = ECCV, year = 2020}

@article{kerbl20233dgs,
  author = {Bernhard Kerbl and Georgios Kopanas and Thomas Leimk{\"u}hler and George Drettakis},
  title = {3D Gaussian Splatting for Real-Time Radiance Field Rendering},
  journal = TOG, volume = 42, number = 4, year = 2023}

@inproceedings{liu2023zero123,
  author = {Ruoshi Liu and Rundi Wu and Basile Van Hoorick and Pavel Tokmakov and Sergey Zakharov and Carl Vondrick},
  title = {Zero-1-to-3: Zero-Shot One Image to 3D Object},
  booktitle = ICCV, year = 2023}

@inproceedings{shi2023mvdream,
  author = {Yichun Shi and Peng Wang and Jianglong Ye and Long Mai and Kejie Li and Xiao Yang},
  title = {MVDream: Multi-View Diffusion for 3D Generation},
  booktitle = ICLR, year = 2024}

@inproceedings{liu2024syncdreamer,
  author = {Yuan Liu and Cheng Lin and Zijiao Zeng and Xiaoxiao Long and Lingjie Liu and Taku Komura and Wenping Wang},
  title = {SyncDreamer: Generating Multiview-Consistent Images from a Single-View Image},
  booktitle = ICLR, year = 2024}

@inproceedings{voleti2024sv3d,
  author = {Vikram Voleti and Chun-Han Yao and Mark Boss and Adam Letts and David Pankratz and Dmitry Tochilkin and others},
  title = {SV3D: Novel Multi-View Synthesis and 3D Generation from a Single Image using Latent Video Diffusion},
  booktitle = ECCV, year = 2024}

@inproceedings{wang2024dust3r,
  author = {Shuzhe Wang and Vincent Leroy and Yohann Cabon and Boris Chidlovskii and J{\'e}r{\^o}me Revaud},
  title = {DUSt3R: Geometric 3D Vision Made Easy},
  booktitle = CVPR, year = 2024}

@inproceedings{wang2025vggt,
  author = {Jianyuan Wang and Minghao Chen and Nikita Karaev and Andrea Vedaldi and Christian Rupprecht and David Novotny},
  title = {VGGT: Visual Geometry Grounded Transformer},
  booktitle = CVPR, year = 2025}

@inproceedings{wang2025pi3,
  author = {Yifan Wang and Jianjun Zhou and Haoyi Zhu and Wenzheng Chang and Yang Zhou and Zizun Li and others},
  title = {$\pi^3$: Permutation-Equivariant Visual Geometry Learning},
  booktitle = ICLR, year = 2026}

@inproceedings{yang2024depthanything,
  author = {Lihe Yang and Bingyi Kang and Zilong Huang and Xiaogang Xu and Jiashi Feng and Hengshuang Zhao},
  title = {Depth Anything: Unleashing the Power of Large-Scale Unlabeled Data},
  booktitle = CVPR, year = 2024}

@inproceedings{lin2025depthanything3,
  author = {Haotong Lin and Sili Chen and Jun Hao Liew and Donny Y. Chen and Zhenyu Li and Guang Shi and Jiashi Feng and Bingyi Kang},
  title = {Depth Anything 3: Recovering the Visual Space from Any Views},
  booktitle = ICLR, year = 2026}

@inproceedings{deitke2023objaverse,
  author = {Matt Deitke and Dustin Schwenk and Jordi Salvador and Luca Weihs and Oscar Michel and Eli VanderBilt and Ludwig Schmidt and Kiana Ehsani and Aniruddha Kembhavi and Ali Farhadi},
  title = {Objaverse: A Universe of Annotated 3D Objects},
  booktitle = CVPR, year = 2023}

@article{li2026mvsam3d,
  author = {Baicheng Li and Dong Wu and Jun Li and Shunkai Zhou and Zecui Zeng and Lusong Li and Hongbin Zha},
  title = {MV-SAM3D: Adaptive Multi-View Fusion for Layout-Aware 3D Generation},
  journal = {arXiv preprint arXiv:2603.11633}, year = 2026}

@inproceedings{ruiz2023dreambooth,
  author = {Nataniel Ruiz and Yuanzhen Li and Varun Jampani and Yael Pritch and Michael Rubinstein and Kfir Aberman},
  title = {DreamBooth: Fine Tuning Text-to-Image Diffusion Models for Subject-Driven Generation},
  booktitle = CVPR, year = 2023}

@inproceedings{gal2023textualinversion,
  author = {Rinon Gal and Yuval Alaluf and Yuval Atzmon and Or Patashnik and Amit H. Bermano and Gal Chechik and Daniel Cohen-Or},
  title = {An Image is Worth One Word: Personalizing Text-to-Image Generation using Textual Inversion},
  booktitle = ICLR, year = 2023}

@inproceedings{kumari2023customdiffusion,
  author = {Nupur Kumari and Bingliang Zhang and Richard Zhang and Eli Shechtman and Jun-Yan Zhu},
  title = {Multi-Concept Customization of Text-to-Image Diffusion},
  booktitle = CVPR, year = 2023}

@inproceedings{hu2022lora,
  author = {Edward J. Hu and Yelong Shen and Phillip Wallis and Zeyuan Allen-Zhu and Yuanzhi Li and Shean Wang and Lu Wang and Weizhu Chen},
  title = {LoRA: Low-Rank Adaptation of Large Language Models},
  booktitle = ICLR, year = 2022}

@article{ye2023ipadapter,
  author = {Hu Ye and Jun Zhang and Sibo Liu and Xiao Han and Wei Yang},
  title = {IP-Adapter: Text Compatible Image Prompt Adapter for Text-to-Image Diffusion Models},
  journal = {arXiv preprint arXiv:2308.06721}, year = 2023}

@inproceedings{tan2025ominicontrol,
  author = {Zhenxiong Tan and Songhua Liu and Xingyi Yang and Qiaochu Xue and Xinchao Wang},
  title = {OminiControl: Minimal and Universal Control for Diffusion Transformer},
  booktitle = ICCV, year = 2025}

@inproceedings{jiang2024videobooth,
  author = {Yuming Jiang and Tianxing Wu and Shuai Yang and Chenyang Si and Dahua Lin and Yu Qiao and Chen Change Loy and Ziwei Liu},
  title = {VideoBooth: Diffusion-based Video Generation with Image Prompts},
  booktitle = CVPR, year = 2024}

@inproceedings{wei2024dreamvideo,
  author = {Yujie Wei and Shiwei Zhang and Zhiwu Qing and Hangjie Yuan and Zhiheng Liu and Yu Liu and Yingya Zhang and Jingren Zhou and Hongming Shan},
  title = {DreamVideo: Composing Your Dream Videos with Customized Subject and Motion},
  booktitle = CVPR, year = 2024}

@article{he2024idanimator,
  author = {Xuanhua He and Quande Liu and Shengju Qian and Xin Wang and Tao Hu and Ke Cao and Keyu Yan and Jie Zhang},
  title = {ID-Animator: Zero-Shot Identity-Preserving Human Video Generation},
  journal = {arXiv preprint arXiv:2404.15275}, year = 2024}

@inproceedings{yuan2025consisid,
  author = {Shenghai Yuan and Jinfa Huang and Xianyi He and Yunyang Ge and Yujun Shi and Liuhan Chen and Jiebo Luo and Li Yuan},
  title = {Identity-Preserving Text-to-Video Generation by Frequency Decomposition},
  booktitle = CVPR, year = 2025}

@inproceedings{liu2025phantom,
  author = {Lijie Liu and Tianxiang Ma and Bingchuan Li and Zhuowei Chen and Jiawei Liu and Gen Li and Siyu Zhou and Qian He and Xinglong Wu},
  title = {Phantom: Subject-Consistent Video Generation via Cross-Modal Alignment},
  booktitle = ICCV, year = 2025}

@inproceedings{deng2025magref,
  author = {Yufan Deng and Xun Guo and Yuanyang Yin and Jacob Zhiyuan Fang and Yiding Yang and Yizhi Wang and others},
  title = {MAGREF: Masked Guidance for Any-Reference Video Generation with Subject Disentanglement},
  booktitle = ICLR, year = 2026}

@inproceedings{jiang2025vace,
  author = {Zeyinzi Jiang and Zhen Han and Chaojie Mao and Jingfeng Zhang and Yulin Pan and Yu Liu},
  title = {VACE: All-in-One Video Creation and Editing},
  booktitle = ICCV, year = 2025}

@inproceedings{chen2025multisubject,
  author = {Tsai-Shien Chen and Aliaksandr Siarohin and Willi Menapace and Yuwei Fang and Kwot Sin Lee and Ivan Skorokhodov and Kfir Aberman and Jun-Yan Zhu and Ming-Hsuan Yang and Sergey Tulyakov},
  title = {Multi-Subject Open-Set Personalization in Video Generation},
  booktitle = CVPR, year = 2025}

@inproceedings{song2026mvs2v,
  author = {Ziyang Song and Xinyu Gong and Bangya Liu and Zelin Zhao},
  title = {MV-S2V: Multi-View Subject-Consistent Video Generation},
  booktitle = {ACM SIGGRAPH Conf. Papers}, year = 2026}

@article{ko2026threedreambooth,
  author = {{Hyun-kyu} Ko and Jihyeon Park and Younghyun Kim and Dongheok Park and Eunbyung Park},
  title = {3DreamBooth: High-Fidelity 3D Subject-Driven Video Generation Model},
  journal = {arXiv preprint arXiv:2603.18524}, year = 2026}

@inproceedings{asim2025met3r,
  author = {Mohammad Asim and Christopher Wewer and Thomas Wimmer and Bernt Schiele and Jan Eric Lenssen},
  title = {{MEt3R}: Measuring Multi-View Consistency in Generated Images},
  booktitle = CVPR, year = 2025}

@inproceedings{huang2024vbench,
  author = {Ziqi Huang and Yinan He and Jiashuo Yu and Fan Zhang and Chenyang Si and Yuming Jiang and others},
  title = {{VBench}: Comprehensive Benchmark Suite for Video Generative Models},
  booktitle = CVPR, year = 2024}

@inproceedings{wang2024internvid,
  author = {Yi Wang and Yinan He and Yizhuo Li and Kunchang Li and Jiashuo Yu and Xin Ma and others},
  title = {{InternVid}: A Large-scale Video-Text Dataset for Multimodal Understanding and Generation},
  booktitle = ICLR, year = 2024}

@article{oquab2024dinov2,
  author = {Maxime Oquab and Timoth\'ee Darcet and Th\'eo Moutakanni and Huy Vo and Marc Szafraniec and Vasil Khalidov and others},
  title = {{DINOv2}: Learning Robust Visual Features without Supervision},
  journal = {Transactions on Machine Learning Research}, year = 2024}

@inproceedings{radford2021clip,
  author = {Alec Radford and Jong Wook Kim and Chris Hallacy and Aditya Ramesh and Gabriel Goh and Sandhini Agarwal and others},
  title = {Learning Transferable Visual Models From Natural Language Supervision},
  booktitle = {ICML}, year = 2021}

@inproceedings{jampani2023navi,
  author = {Varun Jampani and Kevis-Kokitsi Maninis and Andreas Engelhardt and Arjun Karpur and Karen Truong and Kyle Sargent and Stefan Popov and Andre Araujo and Ricardo Martin-Brualla and others},
  title = {{NAVI}: Category-Agnostic Image Collections with High-Quality {3D} Shape and Pose Annotations},
  booktitle = NIPS, year = 2023}

@inproceedings{dong2025dtc,
  author = {Zhao Dong and Ka Chen and Zhaoyang Lv and Hong-Xing Yu and Yunzhi Zhang and Cheng Zhang and Yufeng Zhu and Stephen Tian and Zhengqin Li and Geordie Moffatt and Sean Christofferson and James Fort and Xiaqing Pan and Mingfei Yan and Jiajun Wu and Carl Yuheng Ren and Richard Newcombe},
  title = {Digital Twin Catalog: A Large-Scale Photorealistic {3D} Object Digital Twin Dataset},
  booktitle = {CVPR}, year = 2025}

@inproceedings{downs2022gso,
  author = {Laura Downs and Anthony Francis and Nate Koenig and Brandon Kinman and Ryan Hickman and Krista Reymann and Thomas B. McHugh and Vincent Vanhoucke},
  title = {Google Scanned Objects: A High-Quality Dataset of {3D} Scanned Household Items},
  booktitle = {ICRA}, year = 2022}

@inproceedings{kirillov2023sam,
  title     = {Segment Anything},
  author    = {Kirillov, Alexander and Mintun, Eric and Ravi, Nikhila and Mao, Hanzi and Rolland, Chloe and Gustafson, Laura and Xiao, Tete and Whitehead, Spencer and Berg, Alexander C. and Lo, Wan-Yen and Doll{\'a}r, Piotr and Girshick, Ross},
  booktitle = {ICCV},
  year      = {2023}
}

@inproceedings{liu2024groundingdino,
  title     = {Grounding {DINO}: Marrying {DINO} with Grounded Pre-Training for Open-Set Object Detection},
  author    = {Liu, Shilong and Zeng, Zhaoyang and Ren, Tianhe and Li, Feng and Zhang, Hao and Yang, Jie and Jiang, Qing and Li, Chunyuan and Yang, Jianwei and Su, Hang and Zhu, Jun and Zhang, Lei},
  booktitle = {ECCV},
  year      = {2024}
}

@article{cao2025uni3c,
  title   = {Uni3C: Unifying Precisely {3D}-Enhanced Camera and Human Motion Controls for Video Generation},
  author  = {Cao, Chenjie and Zhou, Jingkai and Li, Shikai and Liang, Jingyun and Yu, Chaohui and Wang, Fan and Xue, Xiangyang and Fu, Yanwei},
  journal = {arXiv:2504.14899},
  year    = {2025}
}
